\documentclass[aos,MSNbibl,nameyear,dvips]{arximspdf}
\input xypic

\doi{10.1214/13-AOS1095} %kopijuoti is PTS
\volume{41}
\issue{3}
\pubyear{2013}
\firstpage{1381}
\lastpage{1405}

\makeatletter
\newcommand{\rrVert}{\Vert}
\newcommand{\llVert}{\Vert}
\newcommand{\eqref}[1]{(\ref{#1})}

\newcommand{\argmin}{\operatorname{argmin}}

\newcommand{\dd}{\mathrm{d}}
\newcommand{\EE}{\mathrm{E}}

\newcommand{\Real}{\mathbb{R}}

\newcommand{\fhat}{\hat{f}}
\newcommand{\fcl}{\check{f}}
\newcommand{\Istar}{I_0}

\newcommand{\calB}{\mathcal{B}}
\newcommand{\calF}{\mathcal{F}}
\newcommand{\calG}{\mathcal{G}}
\newcommand{\calH}{\mathcal{H}}
\newcommand{\calN}{\mathcal{N}}
\newcommand{\calX}{\mathcal{X}}
\newcommand{\boldK}{\mathbf{K}}

\newcommand{\tilb}{b}

\newcommand{\kmin}{\kappa}

\newcommand{\lambdatmp}{{\lambda}}

\newcommand{\boldalpha}{\bolds{\alpha}}

\newcommand{\recond}[1]{\beta_{#1}}

\newcommand{\totH}{\calH}

\newcommand{\sprime}{\tilde{s}}

\newcommand{\Dtr}{D_{\mathrm{tr}}}
\newcommand{\Dte}{D_{\mathrm{te}}}

\newtheorem{Theorem}{Theorem}
\newproclaim{Remark}[Theorem]{Remark}
\newproclaim{Definition}[Theorem]{Definition}
\newtheorem{Lemma}[Theorem]{Lemma}

\newtheorem{Corollary}[Theorem]{Corollary}
\newproclaim{Assumption}{Assumption}

\makeatother

\begin{document}
\begin{frontmatter}

\title{Fast learning rate of multiple kernel learning: Trade-off
between sparsity and smoothness}
\runtitle{Fast learning rate of MKL}

\begin{aug}
\author[a]{\fnms{Taiji} \snm{Suzuki}\corref{}\thanksref{t1}\ead[label=e1]{s-taiji@stat.t.u-tokyo.ac.jp}}
\and
\author[b]{\fnms{Masashi} \snm{Sugiyama}\thanksref{t2}\ead[label=e3]{sugi@cs.titech.ac.jp}}
\runauthor{T. Suzuki and M. Sugiyama}
\affiliation{University of Tokyo and Tokyo Institute of Technology}
\address[a]{
Department of Mathematical Informatics\\
Graduate School of Information Science\\
\quad and Technology\\
University of Tokyo\\
7-3-1 Hongo, Bunkyo-ku\\
Tokyo\\
Japan\\
\printead{e1}}

\address[b]{Department of Computer Science\\
Graduate School of Information Science\\
\quad and Engineering\\
Tokyo Institute of Technology\\
2-12-1 O-okayama, Meguro-ku\\
Tokyo\\
Japan\\
\printead{e3}}
\thankstext{t1}{Supported in part by MEXT KAKENHI 22700289, and the Aihara project, the FIRST program from JSPS, initiated by CSTP.}
\thankstext{t2}{Supported in part by the FIRST program.}
\end{aug}

% HISTORY:
\received{\smonth{12} \syear{2011}}
\revised{\smonth{1} \syear{2013}}

% ABSTRACT
%
\begin{abstract}
We investigate the learning rate of multiple kernel learning (MKL)
with $\ell_1$ and elastic-net regularizations.
The elastic-net regularization is a composition of an $\ell
_1$-regularizer for inducing the sparsity
and an $\ell_2$-regularizer for controlling the smoothness.
We focus on a sparse setting where the total number of kernels is large,
but the number of nonzero components of the ground truth is relatively small,
and show sharper convergence rates
than the learning rates have ever shown for both $\ell_1$ and
elastic-net regularizations.
Our analysis reveals some relations between
the choice of a regularization function and the performance.
If the ground truth is smooth,
we show a faster convergence rate for the elastic-net regularization
with less conditions than $\ell_1$-regularization;
otherwise, a faster convergence rate for the $\ell_1$-regularization
is shown.

%Our analysis shows there appears a trade-off between the sparsity and
%the smoothness
%when it comes to selecting which of $\ell_1$ and elastic-net
%regularizations to use;
%if the ground truth is smooth, the elastic-net regularization is
%preferred,
%otherwise the $\ell_1$ regularization is preferred.

%and prove that $ell_1$ MKL and elastic-net MKL achieve
%the minimax learning rate on the $\ell_1$-mixed-norm ball and $
%Our bound is sharper than the convergence rates ever shown,
%and has a property that the smoother the truth is,
%the faster the convergence rate is.
\end{abstract}

% KEYWORDS

\end{frontmatter}

%s1 #&#
\section{Introduction}
Learning with kernels such as support vector machines has been
demonstrated to
be a promising approach,
given that kernels were chosen appropriately
[\citet{book:Schoelkopf+Smola:2002}, \citet{Book:Taylor+Cristianini:2004}].
% However, since the behavior of the kernel methods relies heavily on
%the choice of kernels,
% the issue of kernel selection needs to be addressed properly for
%making the kernel methods
% has been an important problem.
So far, various strategies have been employed for choosing appropriate kernels,
ranging from simple cross-validation
[\citet{mach:chapelle+vapnik+bousquet:2002}]
to
more sophisticated ``kernel learning'' approaches
[\citet{JMLR:Ong+etal:2005},
\citet{ICML:Argriou+etal:2006},
\citet{NIPS:Bach:2009},
\citet{NIPS:Cortes+etal:nonlinear:2009},
\citet{ICML:Varma+Babu:2009}].

\emph{Multiple kernel learning} (MKL) is one of the systematic approaches
to learning kernels, which tries to find the optimal linear combination
of prefixed base-kernels by convex optimization [\citet
{JMLR:Lanckriet+etal:2004}].
The seminal paper by \citet{ICML:Bach+etal:2004} showed that this
linear-combination
MKL formulation can be interpreted as $\ell_1$-mixed-norm regularization
(i.e., the sum of the norms of the base kernels).
Based on this interpretation,
several variations of MKL were proposed,
and promising performance was achieved by
``intermediate'' regularization strategies between the sparse ($\ell_1$)
and dense ($\ell_2$) regularizers, for example,
a mixture of $\ell_1$-mixed-norm and $\ell_2$-mixed-norm
called the \emph{elastic-net regularization}
[\citet{NIPSWS:Taylor:2008},
\citet{NIPSWS:ElastMKL:2009}]
and $\ell_p$-mixed-norm regularization with $1<p<2$
[\citet{JMLR:MicchelliPontil:2005},
\citet{NIPS:Marius+etal:2009}].

Together with the active development of practical MKL optimization algorithms,
theoretical analysis of MKL has also been extensively conducted.
% While several methods related to MKL have been proposed,
For $\ell_1$-mixed-norm MKL,
\citet{COLT:Koltchinskii:2008} established the learning rate
$d^{{(1-s)}/{(1+s)}} n^{-{1}/{(1+s)}} + {d \log(M)}/{n}$
under rather restrictive conditions,
where $n$ is the number of samples,
$d$ is the number of nonzero components of the ground truth,
$M$ is the number of kernels
and $s$ ($0<s<1$) is a constant representing the complexity
of the reproducing kernel Hilbert spaces (RKHSs).
Their conditions include a smoothness assumption of the ground truth.
% ($q=1$ in our terminology (Assumption~\ref{ass:convolution})).
For elastic-net regularization (which we call \emph{elastic-net MKL}),
\citet{AS:Meier+Geer+Buhlmann:2009} gave a near optimal convergence rate
$d (n/\log(M) )^{-{1}/{(1+s)}}$.
Recently, \citet{AS:Koltchinskii+Yuan:2010}
showed that MKL with a variant of $\ell_1$-mixed-norm regularization
(which we call \emph{$L_1$-MKL}) achieves the minimax optimal
convergence rate,
which successfully captured sharper dependency with respect to $\log(M)$
than the bound of \citet{AS:Meier+Geer+Buhlmann:2009} and
established the bound $d n^{-{1}/{(1+s)}} + {d \log(M)}/{n}$.
Another line of research
considers the cases where the ground truth is not sparse,
and bounds the Rademacher complexity of a candidate kernel class
by a pseudo-dimension of the kernel class
[\citet
{COLT:Srebro+BenDavid:2006},
\citet{COLT:Ying+Campbell:2009},
\citet{UAI:Cortes+etal:2009},
\citet{ECML:Marius+etal:2010}].
Fast learning rate of MKL in nonsparse settings is given
by \citet{JMLR:Kloft+Blanchard:2012} for $\ell_p$-mixed-norm
regularization and
by Suzuki (\citeyear{NIPS:Suzuki:2011,arXiv:Suzuki:MKLUnif:2011}) for
regularizations corresponding to arbitrary monotonically increasing norms.

% However, these works do not utilize the sparsity of the truth,
% and thus are not suited to sparse settings.
%Moreover their order is $1/\sqrt{n}$ with respect to the number of
%samples
%because they do not utilize the {\it localization techniques}.

In this paper, we focus on the sparse setting
(i.e., the total number of kernels is large,
but the number of nonzero components of the ground truth is relatively small),
and derive sharp learning rates for both $L_1$-MKL and elastic-net MKL.
Our new learning rates,
{\makeatletter
\setvaluelist{myfoo}{$L_1$-MKL,Elastic-net MKL}                  %%%pvz.: \setvaluelist{myfoo}{*,**}
\def\theequation{\getitemvalue{myfoo}{\the\c@equation}}
\makeatother
\begin{eqnarray}\hspace*{80pt}
\label{L1MKL}&\displaystyle d^{{(1-s)}/{(1+s)}} n^{-{1}/{(1+s)}}
R_{1,f^*}^{{2s}/{(1+s)}} + \frac{d\log(M)}{n},&\\
\label{ElasticnetMKL} &\displaystyle d^{{(1+q)}/{(1+q+s)}} n^{-
{(1+q)}/{(1+q+s)}} R_{2,g^*}^{{2s}/{(1+q+s)}}
+ \frac{d\log(M)}{n},&
\end{eqnarray}}
\hspace*{-2pt}are faster than all the existing bounds,
where $R_{1,f^*}$ is the $\ell_1$-mixed-norm of the truth,
$R_{2,g^*}$ is a kind of $\ell_2$-mixed-norm of the truth
and $q$ ($0 \leq q \leq1$) is a constant depending on the smoothness
of the ground truth.

% We suppose the ground truth is sparse,
% and the resultant bond is suited to sparse settings.

Our contributions are summarized as follows:
% Our main notion is that,
\begin{longlist}[(a)]
\item[(a)] The sharpest existing bound for $L_1$-MKL given by \citet
{AS:Koltchinskii+Yuan:2010}
achieves the minimax rate on the $\ell_{\infty}$-mixed-norm ball
[Raskutti, Wainwright and Yu
(\citeyear{NIPS:Raskutti+Martin:2009,JMLR:Raskutti+Martin:2012})].
Our work follows this line and shows that the learning rates for
$L_1$-MKL and elastic-net MKL
further achieve the minimax rates on the \emph{$\ell_1$-mixed-norm
ball} and \emph{$\ell_2$-mixed-norm ball}, respectively,
both of which are faster than that on the $\ell_{\infty}$-mixed-norm ball.
This result implies that the bound by \citet{AS:Koltchinskii+Yuan:2010}
is tight
only when the ground truth is evenly spread in the nonzero components.

\item[(b)]
We included the \emph{smoothness} $q$ of the ground truth into our
learning rate,
where the ground truth is said to be smooth
if it is represented as a convolution of a certain function and an
integral kernel; see Assumption~\ref{ass:convolution}.
Intuitively, for larger~$q$, the truth is smoother.
We show that elastic-net MKL properly makes use of the smoothness of
the truth:
The smoother the truth is, the faster the convergence rate of
elastic-net MKL is.
%We show that,
That is, the resultant convergence rate of elastic-net MKL becomes as
if the complexity of RKHSs was $\frac{s}{1+q}$
instead of the true complexity~$s$.
\citet{AS:Meier+Geer+Buhlmann:2009} and \citet
{AS:Koltchinskii+Yuan:2010} assumed $q=0$ and
\citet{COLT:Koltchinskii:2008} considered a situation of $q=1$.
Our analysis covers both of those situations
and is more general since any $0\le q\le1$ is allowed.

\item[(c)]
We investigate a relation between the sparsity and the smoothness. % of
%the estimator.
%According to our analysis,
Roughly speaking, $L_1$-MKL generates a sparser solution while
elastic-net MKL generates a smoother solution.
When the smoothness $q$ of the truth is small (say $q=0$),
we give a faster convergence rate of $L_1$-MKL than that of elastic-net
MKL. %could achieve a faster convergence rate than elastic-net MKL.
On the other hand, if the truth is smooth,
elastic-net MKL can make use of the smoothness of the truth. % with
%less assumptions than $L_1$-MKL.
In that situation, the learning rate of elastic-net MKL could be faster
than $L_1$-MKL.
\end{longlist}
% We show that there is a trade-off between the sparsity and the
%smoothness of the estimator.
% While $L_1$-MKL gives a sparser solution than elastic-net,
% a smoother solution is generated by elastic-net MKL.
% Our analysis claims that when the smoothness $q$ of the truth is
%small (say $q=0$),
% $L_1$-MKL achieves a faster convergence rate than elastic-net MKL.
% On the other hand, if the truth is smooth,
% the learning rate of elastic-net MKL could be faster than $L_1$-MKL.

%t1 #&#
%
\begin{table}[b]
\tabcolsep=0pt
\caption{Relation between our analysis and existing analyses}
\label{tab:RelationOfBounds}
\begin{tabular*}{\textwidth}{@{\extracolsep{\fill}}lcccc@{}}
\hline
& \textbf{Penalty} & \textbf{Smoothness} & \textbf{Minimax} & \multicolumn{1}{c@{}}{\textbf{Convergence rate}} \\
& & \multicolumn{1}{c}{$\bolds{(q)}$} & \textbf{optimality} & \\
\hline
KY (2008)
%K\&Y (2008) \citet{COLT:Koltchinskii:2008}
& $\ell_1$ & $q=1$ & ? & $d^{{(1-s)}/{(1+s)}}n^{-{1}/{(1+s)}} +
\frac{d\log(M)}{n}$ \\[2pt]
MGB (2009)
%M,G\&B (2009) \citet{AS:Meier+Geer+Buhlmann:2009}
& elastic-net & $q=0$ & $\times$ & $ (\frac{\log(M)}{n}
)^{{1}/{(1+s)}}(d+R_{2,g^*}^2)$ \\[2pt]
KY (2010)
%K\&Y (2010) \citet{AS:Koltchinskii+Yuan:2010}
& $\ell_1$ & $q=0$ & $\ell_\infty$-ball & $\frac{(d + R_{1,f^*
})}{n^{{1}/{(1+s)}}} + \frac{d\log(M)}{n}$ \\[6pt]
This paper & elastic-net & $0\leq q\leq1$ & $\ell_2$-ball & $
(\frac{d}{n} )^{{(1+q)}/{(1+q+s)}} R_{2,g^*}^{
{2s}/{(1+q+s)}} + \frac{d\log(M)}{n}$ \\[2pt]
& $\ell_1$ & $q=0$ & $\ell_1$-ball & $\frac{d^{
{(1-s)}/{(1+s)}}}{n^{{1}/{(1+s)}}}R_{1,f^*}^{{2s}/{(1+s)}} + \frac
{d\log(M)}{n}$ \\
\hline
\end{tabular*}
\end{table}

The relation between our analysis and existing analyses is summarized
in Table~\ref{tab:RelationOfBounds}.

%s2 #&#
\section{Preliminaries}
\label{sec:Preliminary}
In this section, we formulate elastic-net MKL,
and summarize mathematical tools that are needed for our theoretical analysis.

%s2.1 #&#
\subsection{Formulation}

Suppose we are given $n$ samples $\{(x_i,y_i)\}_{i=1}^n$ where $x_i$
belongs to an input space $\calX$ and $y_i \in\Real$.
We denote the marginal distribution of $X$ by~$\Pi$.
We consider an MKL regression problem
in which the unknown target function is represented
as $f(x)= \sum_{m=1}^M f_m(x)$,
where each $f_m$ belongs to a different RKHS
$\calH_m (m = 1,\ldots,M)$ with a kernel $k_m$ over $\calX\times
\calX$.

The elastic-net MKL we consider in this paper is the version considered
in \citet{AS:Meier+Geer+Buhlmann:2009},
%
%e1 #&#
%
\renewcommand{\theequation}{\arabic{equation}}
\setcounter{equation}{0}
\begin{eqnarray}
\label{eq:primalElasticMKLnonpara} \fhat&= &\mathop{\arg\min}_{f_m
\in\calH_m \atop(m=1,\ldots,M)}
\frac{1}{n}\sum_{i=1}^N
\Biggl(y_i - \sum_{m=1}^M
f_m(x_i) \Biggr)^{ 2}
\nonumber
\\[-8pt]
\\[-8pt]
\nonumber
&&{} + \sum_{m=1}^M \bigl( {\lambda_1^{(n)}} \|f_m\|_n + {\lambda
_2^{(n)}} \|f_m \|_{\calH_{m}} + {\lambda_3^{(n)}}\|f_m
\|_{\calH_{m}}^2 \bigr),
\end{eqnarray}
where $\|f_m\|_n:= \sqrt{ \frac{1}{n} \sum_{i=1}^n f_m(x_i)^2}$ and
$\|f_m\|_{\calH_{m}}$ is the RKHS norm of $f_m$ in~$\calH_m$.
The regularizer is the mixture of $\ell_1$-term $\sum_{m=1}^M
({\lambda_1^{(n)}} \|f_m\|_n +\break {\lambda_2^{(n)}} \|f_m \|_{\calH
_{m}})$ and $\ell_2$-term $\sum_{m=1}^M
{\lambda_3^{(n)}}\|f_m\|_{\calH_{m}}^2$.
In that sense, we say that the regularizer is of the elastic-net
type\setcounter{footnote}{2}\footnote{%
There is another version of MKL with elastic-net regularization
considered in \citet{NIPSWS:Taylor:2008} and \citet{NIPSWS:ElastMKL:2009},
that is,
$ %\begin{eqnarray}
{\lambda_2^{(n)}}\sum_{m=1}^M \|f_m\|_{\calH_{m}} + {\lambda
_3^{(n)}}\sum_{m=1}^M \|
f_m\|_{\calH_{m}}^2
$ %\end{eqnarray}
(i.e., there is no $\|f_m\|_n$ term in the regularizer).
However, we focus on equation (\ref{eq:primalElasticMKLnonpara}) because
the above one is too loose to properly bound the irrelevant components
of the estimated function. % and the truth on the irrelevant components.
%that is more preferable in sparse settings from theoretical point of
%view.
%In fact, our formulation achieves the minimax optimal rate for sparse
%ground truth.
} [\citet{JRSS:Zou+Hastie:2005}].
Here the $\ell_1$-term is a mixture of the empirical $L_2$-norm $\|
f_m\|_n$ and the RKHS norm $\|f_m\|_{\calH_{m}}$.
\citet{AS:Koltchinskii+Yuan:2010} considered $\ell_1$-regularization
that contains only the $\ell_1$-term:
$\sum_m ({\lambda_1^{(n)}}\|f_m\|_n + {\lambda_2^{(n)}}\|f_m\|
_{\calH_{m}})$.
To distinguish the situations of ${\lambda_3^{(n)}}= 0$ and ${\lambda
_3^{(n)}}>
0$, we refer to
the learning method \eqref{eq:primalElasticMKLnonpara} with
${\lambda_3^{(n)}}= 0$ as \textit{$L_1$-MKL} and that with ${\lambda_3^{(n)}}
>0$ as \textit{elastic-net MKL}.

By the representer theorem [\citet{JMAA:KimeldorfWahba:1971}],
the solution $\fhat$ can be expressed as a linear combination of $nM$ kernels:
$\exists\alpha_{m,i}\in\Real, \fhat_m(x) = \sum_{i=1}^n \alpha
_{m,i}k_m(x,x_i)$.
Thus, using the Gram matrix $\boldK_m = (k_m(x_i,x_j))_{i,j}$,
the regularizer in \eqref{eq:primalElasticMKLnonpara} is expressed as
\[
\sum_{m=1}^M \biggl( {\lambda_1^{(n)}}\sqrt
{ \boldalpha_m^\top\frac{\boldK_m \boldK_m}{n}
\boldalpha_m} +{\lambda_2^{(n)}}\sqrt{ \boldalpha_m^\top
\boldK_m \boldalpha_m} %\sqrt{ \boldalpha_m^\top\left(\frac{\boldK_m
+
{\lambda_3^{(n)}}%\sum_{m=1}^M
\boldalpha_m^\top
\boldK_m \boldalpha_m \biggr),
\]
where $\boldalpha_m = (\alpha_{m,i})_{i=1}^n \in\Real^n$.
Thus, we can solve the problem by an SOCP (second-order cone
programming) solver as in \citet{ICML:Bach+etal:2004},
the coordinate descent algorithms [\citet{JRSS:Meier+etal:2008}] or
the alternating direction method of multipliers [\citet{FTML:Boyd+etal:2011}].

%s2.2 #&#
\subsection{Notation and assumptions}

Here, we present several assumptions used in our theoretical analysis
and prepare notation.

Let $\totH= \calH_1 \oplus\cdots\oplus\calH_M$. %= \{\sum_{m=1}^M
%f_m \mid f_m \in\calH_m \}$.
We utilize the same notation $f \in\totH$ indicating both the vector
$(f_1,\ldots,f_M)$ and the function $f = \sum_{m=1}^M f_m$ ($f_m \in
\calH_m$).
This is a little abuse of notation because the decomposition $f = \sum
_{m=1}^M f_m$ might not be unique as an element of $L_2(\Pi)$.
However, this will not cause any confusion.
We denote by $f^*\in\totH$ the ground truth satisfying the
following assumption
(the decomposition $f^*= \sum_{m=1}^M f^*_m$ of the truth
might not be unique but we fix one possibility).

%as1 #&#
%
\begin{Assumption}[(Basic assumptions)]
\label{ass:basic}
\begin{longlist}[(A\ref{ass:basic}-2)]
\item[{(A\ref{ass:basic}-1)}] %$\mathrm{(1)}$]
There exists $f^*= (f^*_1,\ldots,f^*_M) \in
\totH$
such that $\EE[Y|X] = \break\sum_{m=1}^M f^*_m(X)$,
and the noise $\varepsilon_i:= y_i - f^*(x_i)$
%has a strictly positive variance;
%there exists $\sigma>0$ such that $\EE[\varepsilon^2 | X] > \sigma^2 $
%for all $X \in\calX$.
%We also assume that $\varepsilon$
is bounded as $|\varepsilon_i| \leq L$ (a.s.).
\item[(A\ref{ass:basic}-2)]
For each $m=1,\ldots,M$, %$\calH_m$ is separable,
the kernel function $k_m$ is continuous
and $\sup_{X\in\calX} |k_m(X,X)| \leq1$.
\end{longlist}
\end{Assumption}
%
%assumption (A1)?============}
The first assumption in (A\ref{ass:basic}-1) ensures the model $\calH
$ is correctly specified,
and the technical assumption $|\varepsilon_i| < L$ allows $\varepsilon
_i f$
to be Lipschitz continuous with respect to $f$.
The assumption of correct specification can be relaxed to misspecified
settings, and
the bounded noise can be replaced with i.i.d. Gaussian noise as in
\citet{JMLR:Raskutti+Martin:2012}.
% \citet{JMLR:BachConsistency:2008,FCM:Caponetto+Vito:2007}, but
However, for the sake of simplicity,
we assume these conditions.
It is known that assumption (A\ref{ass:basic}-2) gives the relation
$\|f_m\|_{\infty} \leq\|f_m\|_{\calH_{m}}$; see Chapter~4 of \citet
{Book:Steinwart:2008}. % between the $\ell_\infty$-norm and the RKHS
%norm:

%It is known that the assumption (A\ref{ass:basic}-2) gives the
%following relation
%(see Chapter 4 of \citet{Book:Steinwart:2008}): % between the $\ell_
%%&
% \leq\sup_{x} \| k_m(x,\cdot)\hnorm{m} \|f_m\hnorm{m} %\\
%%&
% \leq
% \leq\|f_m\hnorm{m}.
%%Later, we will also assume a stronger (but practical) condition on
%the sup-norm in Assumption~\ref{ass:supnorm}.
%% The assumption (A3) ensures that $\fstar$ is unique.

%$M$ & The number of candidate kernels. \\ \hline
%$d$ & The number of active kernels of the truth; i.e., $d=|I_0|$. \\
%$R$ & The upper bound of $\sum_{m=1}^M (\|\fstar_m \hnorm{m} +
% \|\fstar_m \hnorm{m}^2)$; see (A4). \\ \hline
%$s$ & The spectral decay coefficient; see (A5). \\ \hline

%We define an operator $\Sigma_{m,m'}:\calH_{m'} \to\calH_{m}$ as
%$$
%$$
%In particular, we denote $\Sigma_{m,m}$ by $T_m$:
Let an integral operator $T_m\dvtx L_2(\Pi)\to L_2(\Pi)$
corresponding to a kernel
function $k_m$ be
\[
T_m f = \int k_m(\cdot,x) f(x) \,\dd\Pi(x).
\]
It is known that this operator is compact, positive and self-adjoint
[see Theorem~4.27 of \citet{Book:Steinwart:2008}],
and hence the spectral theorem shows that there exist an at most countable
orthonormal system $\{\phi_{\ell,m}\}_{\ell=1}^\infty$ and
eigenvalues $\{\mu_{\ell,m}\}_{\ell=1}^\infty$
such that
%
%e2 #&#
%
\begin{equation}
T_m f = \sum_{\ell=1}^\infty
\mu_{\ell,m} \langle\phi_{\ell,m}, f \rangle_{L_2(\Pi)}
\phi_{\ell,m} \label{eq:spectralRepre}
\end{equation}
for $f \in L_2(\Pi)$.
Here we assume $\{\mu_{\ell,m}\}_{\ell=1}^\infty$ is sorted in
descending order,
that is,
$\mu_{1,m} \geq\mu_{2,m} \geq\mu_{3,m} \geq\cdots\geq0$.\vadjust{\goodbreak}
%Using the inclusion map $\iota:\calH_m \to\LPi$, we can the operator
%$T_{k_m}$ can be interpreted as $T_{k_m}: \calH_m \to\calH_m$
Associated with $T_m$, we can define an operator $\tilde
{T}_m\dvtx\calH_m \to\calH_m$ as
\[
\bigl\langle f'_m, \tilde{T}_m
f_m \bigr\rangle_{\calH_m} = \EE\bigl[f_m'(X)f_m(X)
\bigr] = \biggl\langle f_m', \int k_m(
\cdot,x) f_m(x) \,\dd\Pi(x) \biggr\rangle_{\calH_m}.
\]
For the canonical inclusion map $\iota_m\dvtx\calH_m \to L_2(\Pi)$,
one can
check that the following commutative relation holds:
\newcommand{\LPi}{L_2(\Pi)}
\begin{eqnarray*}
&\displaystyle T_m \iota_m f_m = \iota_m \tilde{T}_m f_m, \\
&
\displaystyle
\xymatrix{
\calH_m \ar[d]_{\iota_m}\ar[r]^{\tilde{T}_m }& \calH_m \ar[d]^{\iota_m } \\
\LPi \ar[r]^{T_m } &\LPi.}&
\end{eqnarray*}
Thus we use the same notation for $T_m$ and $\tilde{T}_m$ and
denote by $T_m$ referring to both operators.

Due to Mercer's theorem [\citet{IO:Ferreira+Menegatto:2009}],
$k_m$ has the following spectral expansion:
\[
k_m\bigl(x,x'\bigr) = \sum
_{k=1}^{\infty} \mu_{k,m} \phi_{k,m}(x)
\phi_{k,m}\bigl(x'\bigr), %\label{eq:spectralRepre}
\]
where the convergence is absolute and uniform.
Thus, the inner product of the RKHS $\calH_m$ can be expressed as
$
\langle f_m,g_m \rangle_{\calH_m} = \sum_{k=1}^{\infty} \mu
_{k,m}^{-1} \langle f_m, \phi_{k,m} \rangle_{L_2(\Pi)} \times \langle\phi
_{k,m}, g_m \rangle_{L_2(\Pi)}.
$

%there are an orthonormal system $\{\phi_{k,m}\}_{k,m}$ in $L_2(\Pi)$
%and the spectrum $\{\mu_{k,m}\}_{k,m}$
%such that $k_m$ has the following spectral representation: %is
%represented as
%k_m(x,x') = \sum_{k=1}^{\infty} \mu_{k,m} \phi_{k,m}(x) \phi_{k,m}(x').
%By this spectral representation, the inner product of the RKHS $
%$
%$

The following assumption is regarding the smoothness of the true
function~$f^*_m$.
%
%as2 #&#
%
\begin{Assumption}[(Convolution assumption)]
\label{ass:convolution}
There exist a real number $0 \leq q \leq1$ and $g^*_m \in\calH
_m$ such that
% (\forall m = 1,\ldots,M), && %\label{eq:fstarSigmacond}
%where $k_m^{(q/2)}(x,x')= \sum_{k=1}^{\infty} \mu_{k,m}^{q/2}
% \phi_{k,m}(x) \phi_{k,m}(x')$.
%This is equivalent to the following operator representation:
%
\renewcommand{\theequation}{A\arabic{Assumption}}
\begin{equation}
 f^*_m = T_m^{{q}/{2}}
g^*_m.
\end{equation}
\end{Assumption}
We denote $(g^*_1,\ldots,g^*_M)$ and $\sum_{m=1}^M g^*_m$
by $g^*$
(we use the same notation for both ``vector'' and ``function''
representations with a slight abuse of notation).
The constant $q$ represents the smoothness of the truth $f^*_m$
because $f^*_m$ is generated by operating the integral operator
$T_m^{{q}/{2}}$ to $g^*_m$
($f^*_m(x) = \sum_{\ell=1}^\infty\mu_{\ell,m}^{{q}/{2}}
\langle\phi_{\ell,m}, g^*_m \rangle_{L_2(\Pi)} \times \phi_{\ell,m}(x) $),
and high-frequency components are suppressed as $q$ becomes large.
Therefore, as $q$ becomes larger, $f^*$ becomes ``smoother.''
Assumption (A\ref{ass:convolution}) was considered in \citet
{FCM:Caponetto+Vito:2007} to analyze
the convergence rate of least-squares estimators in a single kernel setting.
In MKL settings,
\citet{COLT:Koltchinskii:2008} showed a fast learning rate of MKL\vadjust{\goodbreak}
assuming $q=1$,
and \citet{JMLR:BachConsistency:2008} showed the consistency of MKL
under $q=1$.
%It ensures the consistency of the least-squares estimates in terms of
%the RKHS norm.
%This condition (for $q=1$) was also assumed in
Proposition 9 of \citet{JMLR:BachConsistency:2008} gave
a sufficient condition to fulfill (A\ref{ass:convolution}) with $q=1$
for translation invariant kernels $k_m(x,x') = h_m(x-x')$.
\citet{AS:Meier+Geer+Buhlmann:2009} considered a situation with $q=0$
on Sobolev space;
the analysis of \citet{AS:Koltchinskii+Yuan:2010} also corresponds to $q=0$.
Note that (A\ref{ass:convolution}) with $q=0$
imposes nothing on the smoothness about the truth,
and our analysis also covers this case.

We show in Appendix~\ref{appendix:CoveringNumber} that
as $q$ increases, the space of the functions that satisfy (A\ref
{ass:convolution}) becomes ``simpler.''
Thus, it might be natural to expect that, under convolution assumption
(A\ref{ass:convolution}),
the learning rate becomes faster as $q$ increases.
Although this conjecture is actually true,
it is not obvious because the convolution assumption only restricts the
ground truth,
not the search space.

%(Technically, $\|\fstar_m \hnorm{m} - \|f_m\hnorm{m} \leq\| \fstar_m
%- f_m \|_{\LPi} \frac{\|\gstar_m\hnorm{m}}{\|\fstar_m\hnorm{m}}$
%(see \eqref{eq:secondboundforbasic}) gives a sharp convergence bound)
%Using the spectral representation \eqref{eq:spectralRepre},
%the condition $\gstar_m \in\calH_m$ is expressed as

%There exists a constant $R$ such that
%for all $m$.

Next we introduce a parameter representing the complexity of RKHSs.
By Theorem 4.27 of \citet{Book:Steinwart:2008}, the sum of $\mu_{\ell,m}$ is bounded ($\sum_{\ell} \mu_{\ell,m} < \infty$),
and thus $\mu_{\ell,m}$ decreases with order $\ell^{-1}$ ($\mu
_{\ell,m} = o(\ell^{-1})$).
We further assume the sequence of the eigenvalues converges even faster
to zero.

%as3 #&#
%
\begin{Assumption}[(Spectral assumption)]
\label{ass:specass}
There exist $0 < s < 1$ and $c$ such that %the spectrum $\mu_{k,m}$ has
%the following decreasing exponent
%
\renewcommand{\theequation}{A\arabic{Assumption}}
\begin{equation}
\mu_{j,m} \leq c
j^{-{1}/{s}},\qquad (1\leq\forall j, 1\leq\forall m \leq M),
\end{equation}
where $\{\mu_{j,m}\}_{j=1}^\infty$ is the spectrum of the kernel
$k_m$; see equation \eqref{eq:spectralRepre}.
\end{Assumption}

It was shown that spectral assumption (A\ref{ass:specass})
gives a bound on the \textit{entropy number} of the RKHSs [\citet
{COLT:Steinwart+etal:2009}].
Remember that
the $\varepsilon$-covering number $\calN(\varepsilon,\mathcal
{B}_{\calG
},L_2(\Pi))$ with respect to $L_2(\Pi)$
for a Hilbert space $\calG$
is the minimal number of balls with radius $\varepsilon$ needed to cover
the unit ball $\mathcal{B}_{\calG}$ in~$\calG$ [\citet
{Book:VanDerVaart:WeakConvergence}].
The $i$th entropy number $e_i(\calG\to L_2(\Pi))$ is the infimum of
$\varepsilon> 0$ for which
$
\calN(\varepsilon,\calB_{\calG},L_2(\Pi)) \leq2^{i-1}.
$
%where $\calB_{\calG}$ is the unit ball of $\calG$ and $\calN(
%of $
%}.
If spectral assumption~(A\ref{ass:specass}) holds, there exists a
constant $\tilde{c}$ that
depends only on $s$ and $c$ such that
the $i$th entropy number is bounded as
%
%e3 #&#
%
\renewcommand{\theequation}{\arabic{equation}}
\setcounter{equation}{2}
\begin{equation}
\label{eq:entropycondition} e_i\bigl(\calH_m \to L_2(\Pi)\bigr) \leq
\tilde{c} i^{- {1}/{(2s)}}, %\label{eq:coveringcondition}
\end{equation}
and the converse is also true; see Theorem 15 of \citet
{COLT:Steinwart+etal:2009} and \citet{Book:Steinwart:2008} for details.
%and the converse is also true (see Theorem 15 of
%details).
Therefore,
if $s$ is large,
at least one of the RKHSs is ``complex,''
and if $s$ is small, all the RKHSs are ``simple.''
A~more detailed characterization of the entropy number
in terms of the spectrum is provided in Appendix~\ref{appendix:CoveringNumber}.
The entropy number of the space of functions that satisfy
the Convolution assumption (A\ref{ass:convolution}) is also provided there.

Finally, we impose the following technical assumption related to the
sup-norm of members in the RKHSs.\vadjust{\goodbreak}
%
%as4 #&#
%
\begin{Assumption}[(Sup-norm assumption)]
\label{ass:supnorm}
Along with the spectral assumption~(A\ref{ass:specass}),
there exists a constant $C_1$ such that
\renewcommand{\theequation}{A\arabic{Assumption}}
\begin{equation}
\|f_m\|_{\infty} \leq
C_1 \|f_m\|_{L_2(\Pi)}^{1-s}
\|f_m\|_{\calH_{m}}^s \qquad(\forall f_m \in
\calH_m,m=1,\ldots,M),
\end{equation}
where $s$ is the exponent defined in spectral assumption (A\ref{ass:specass}).
\end{Assumption}
This assumption might look a bit strong, but this is satisfied if the
RKHS is a Sobolev space or
is continuously embeddable in a Sobolev space.
For example, the RKHSs of Gaussian kernels are continuously embedded in
all Sobolev spaces,
and thus satisfy sup-norm assumption (A\ref{ass:supnorm}).
More generally, RKHSs with $\gamma$-times continuously differentiable
kernels on a closed Euclidean ball in $\Real^d$ are also
continuously embedded in a Sobolev space, and satisfy the sup-norm
assumption (A\ref{ass:supnorm}) with $s=\frac{d}{2\gamma}$; see
Corollary 4.36 of \citet{Book:Steinwart:2008}.
Therefore, this assumption is common for practically used kernels.
A more general necessary and sufficient condition in terms of \textit
{real interpolation} is shown in \citet{Book:Bennett+Sharpley:88}.
\citet{COLT:Steinwart+etal:2009} used this assumption to show the
optimal convergence rates for regularized regression with a single
kernel function where the true function is not contained in the model,
and
one can find detailed discussions about the assumption there.

%Similarly, we define $\Sigma_{I,J}$ and $\VCor_{I,J}$ as the
%restriction to an index set $I \times J$ for $I,J \subseteq\{1,\ldots,M
We denote by $\Istar$ the indices of truly active kernels, that is,
\[
\Istar:=\bigl\{m \mid\bigl\|f^*_m\bigr\|_{\calH_{m}}>0\bigr\}.
\]
%
%and define the complement of $\Istar$ as $\Jstar= {\Istar}^c$.
We define the number of truly active components as $d:= |\Istar|$. For
$f =\sum_{m=1}^M f_m \in\totH$
and a subset of indices $I \subseteq\{1,\ldots,M\}$, we define $\calH
_I =\break \bigoplus_{m\in I} \calH_m$, and
denote by $f_I \in\calH_I$ the restriction of $f$ to an index set
$I$, that is, $f_I = \sum_{m \in I} f_m$.

Now we introduce a geometric quantity that represents dependency
between RKHSs.
That quantity is related to the restricted eigenvalue condition [\citet
{AS:Bickel+etal:2009}]
and is required to show a nice convergence property of MKL.
For a given set of indices $I \subseteq\{1,\ldots, M \}$
and $\tilb\geq0$, we define
\begin{eqnarray*}
&&\recond{\tilb}(I):= \sup\biggl\{ \recond{} >0 \Big| \beta\leq\frac
{\| \sum_{m=1}^M f_m\|_{L_2(\Pi)}} {
( \sum_{m \in I} \| f_m\|_{L_2(\Pi)}^2 )^{{1}/{2}}},\\
&&\hspace*{65pt} \forall f\in\totH\mbox{ such that } \tilb\sum_{m\in I}
\|f_m\|_{L_2(\Pi)} \geq\sum_{m\notin I}
\|f_m\| _{L_2(\Pi)} \biggr\}.
\end{eqnarray*}
For $I=I_0$, we abbreviate $\recond{b}(I_0)$ as
\[
\recond{b}:= \recond{b}(I_0).
\]
This quantity plays an important role in our analysis.
Roughly speaking, this represents the correlation between RKHSs under the
condition that the components within the relevant indices $I$ well ``dominate''
the rest of the components.
One can see that $\recond{\tilb}(I)$ is nonincreasing with respect to
$\tilb$.
The quantity $\recond{\tilb}$ is first introduced by \citet
{AS:Bickel+etal:2009} to define the restricted eigenvalue condition
in the context of parametric model such as the Lasso and the Dantzig selector.
In the context of MKL, \citet{AS:Koltchinskii+Yuan:2010} introduced
this quantity to analyze a convergence rate of $L_1$-MKL.
%Thus, by the incoherence condition and Lemma~\ref{lem:incoherence_ineq},
We will assume that $\recond{\tilb}(I_0)$ is bounded from below with
some $\tilb> 0$
so that
we may focus on bounding the $L_2(\Pi)$-norm of the
``low-dimensional'' components
$\{\fhat_m - f^*_m\}_{m\in I_0}$, instead of all the components.

Here we give a sufficient condition that $\recond{\tilb}(I)$ is
bounded from below.
For a given set of indices $I \subseteq\{1,\ldots, M \}$, % and $J =
%I^c$,
we introduce a quantity $\kappa(I)$ representing the correlation of
RKHSs inside the
indices $I$,
\[
\kmin(I):= \sup\biggl\{\kappa\geq0 \Big| \kappa\leq
\frac{\|\sum_{m\in I}f_m\|_{L_2(\Pi)}^2}{\sum_{m\in I}\|f_m\|
_{L_2(\Pi)
}^2}, \forall f_m \in\calH_m\ (m\in I)
\biggr\}. %,
\]
Similarly, we define the \textit{canonical correlations} of RKHSs
between $I$ and $I^c$ as follows:
\[
\rho(I):= \sup\biggl\{\frac{\langle f_I, g_{I^c} \rangle_{L_2(\Pi)} }{\|
f_I\|
_{L_2(\Pi)}\|g_{I^c} \|_{L_2(\Pi)}} \Big| f_I \in
\calH_I, g_{I^c} \in\calH_{I^c}, f_I
\neq0, g_{I^c} \neq0 \biggr\}.
\]
These quantities give a connection between the $L_2(\Pi)$-norm of $f
\in
\totH$ and the $L_2(\Pi)$-norm of $\{f_m\}_{m\in I}$
as shown in the following lemma. The proof is given in Appendix \ref
{sec:appendixLemm}.
%
%le1 #&#
%
\begin{Lemma}
\label{lem:incoherence_ineq}
For all $I \subseteq\{1,\ldots,M\}$, we have
\[
\| f \|_{L_2(\Pi)}^2 \geq\bigl(1- \rho(I)^2\bigr)
\kmin(I) \biggl(\sum_{m \in
I}\| f_m
\|_{L_2(\Pi)}^2 \biggr), %\label{eq:incoherenceIneq}
\]
thus
\[
\recond{\infty}(I) \geq\sqrt{\bigl(1- \rho(I)^2\bigr) \kmin(I)}.
\]
\end{Lemma}
\citet{COLT:Koltchinskii:2008} and \citet{AS:Meier+Geer+Buhlmann:2009}
analyzed statistical properties of MKL
under the \textit{incoherence condition} where $(1- \rho(I_0)^2)
\kmin(I_0)$ is bounded from below,
that is, RKHSs are not too dependent on each other.
In this paper, we employ a less restrictive condition where $\recond
{\tilb}$ is bounded from below for some positive real $\tilb$.

\section{Convergence rate analysis}
\label{sec:ConvRateAnalysis}
In this section, we present our main result.

%s3.1 #&#
\subsection{The convergence rate of $L_1$-MKL and elastic-net MKL}
Here we derive the learning rate of the estimator $\fhat$
defined by equation (\ref{eq:primalElasticMKLnonpara}).
%We denote the number of truly active components by $d:= |\Istar|$.
We may suppose that the number\vadjust{\goodbreak} of kernels $M$ and
the number of active kernels $d$ are increasing
with respect to the number of samples $n$.
Our main purpose of this section is to show that the learning rate can be
faster than the existing bounds.
The existing bound has already been shown to be optimal on the $\ell
_{\infty}$-mixed-norm ball [\citet
{AS:Koltchinskii+Yuan:2010},
\citet{JMLR:Raskutti+Martin:2012}].
Our claim is that the convergence rates can further achieve
the minimax optimal rates on the \textit{$\ell_1$-mixed-norm ball}
and \emph{$\ell_2$-mixed-norm ball},
which are faster than that on the $\ell_{\infty}$-mixed-norm ball.

%First we derive the learning rate.

Define $\eta(t)$ for $t>0$ and $\xi_n(\lambdatmp)$ for given
$\lambdatmp> 0$ as
\[
\eta(t):= \max(1,\sqrt{t},t/\sqrt{n}),\quad % \xi_n:= \xi_n(
\xi_n:= \xi_n(\lambdatmp) = \max\biggl(
\frac{\lambdatmp^{-
{s}/{2}}}{\sqrt{n}}, \frac{\lambdatmp^{-{1}/{2}}}{n^{
{1}/{(1+s)}}}, \sqrt{\frac{\log(M)}{n}} \biggr).
\]
For a given function $f = \sum_{m=1}^M f_m \in\calH$ and $1\leq p
\leq\infty$, we define the $\ell_p$-mixed-norm of $f$ as
\[
R_{p,f}:= \Biggl(\sum_{m=1}^M
\|f_m \|_{\calH_{m}}^p \Biggr)^{{1}/{p}}.
\]
Let
\[
\tilb_1 = 16 \biggl(1 + \frac{ \sqrt{d} \max_{m\in I_0} \|g^*
_m \|_{\calH_{m}} }{R_{2,g^*}} \biggr),\qquad
\tilb_2 = 16.
\]
Then we obtain the convergence rate of $L_1$- and elastic-net MKL as follows.

%th2 #&#
%
\begin{Theorem}[(Convergence rate of $L_1$-MKL and elastic-net MKL)]
\label{th:TheConvergenceRateMain}
Suppose Assumptions~\ref{ass:basic}--\ref{ass:supnorm} are satisfied.
Then there exist constants $\tilde{C}_1,\tilde{C}_2$ and $\psi_s$
depending on $s,c,L,C_1$
such that
the following convergence rates hold:

{(\ref{ElasticnetMKL})}.
Set ${\lambda_1^{(n)}}= \psi_s \eta(t) \xi_n(\lambdatmp)$,
${\lambda_2^{(n)}}=
{\lambda_1^{(n)}}\lambdatmp^{{1}/{2}}$,
${\lambda_3^{(n)}}= \lambdatmp$ where
$
\lambdatmp= d^{{1}/{(1+q+s)}} n^{-{1}/{(1+q+s)}}R_{2,g^*
}^{-{2}/{(1+q+s)}}.
$
Then
for all $n$ satisfying $\frac{\log(M)}{\sqrt{n}} \leq1$ and
%
%e4 #&#
%
\renewcommand{\theequation}{\arabic{equation}}
\setcounter{equation}{3}
\begin{equation}
\label{eq:recondC1bound} \frac{\tilde{C}_1}{\recond{\tilb_1}^2}
\psi_s \sqrt{n} \xi
_n(\lambdatmp)^2 d \leq1,
\end{equation}
the generalization error of elastic-net MKL is bounded as
%
%e5 #&#
%
\begin{eqnarray}
\label{eq:Elastbound} &&\bigl\|\fhat- f^*\bigr\|_{L_2(\Pi)}^2
\nonumber\\[-2pt]
&&\qquad\leq \frac{\tilde{C}_2}{\recond{\tilb_1}^2} \biggl( d^{
{(1+q)}/{(1+q+s)}}n^{-{(1+q)}/{(1+q+s)}}R_{2,g^*}^{{2s}/{(1+q+s)}}
\nonumber
\\[-9pt]
\\[-9pt]
\nonumber
&&\hspace*{33pt}\qquad{}+ d^{{(q+s)}/{(1+q+s)}} n^{-{(1+q)}/{(1+q+s)} -
{q(1-s)}/{((1+s)(1+q+s))}}
\\[-2pt]
&&\hspace*{158pt}\qquad{}\times R_{2,g^*}^{{2}/{(1+q+s)}}+ \frac{d \log(M)}{n} \biggr) \eta(t)^2,
\nonumber
\end{eqnarray}
with probability
$1- \exp(- t) -
\exp(-\min\{\frac{\recond{\tilb_1}^4 \log(M)}{\tilde
{C}_1^2 \psi_s^2 n\xi_n(\lambdatmp)^4 d^2},
\frac{\recond{\tilb_1}^2}{\tilde{C}_1 \psi_s \xi_n(\lambdatmp)^2
d} \} )$ for all \mbox{$t \geq1$}.\vadjust{\goodbreak}
%$1- \exp(- t) - \exp[-\min\left\{\log(M)/(n\xi_n(\lambdatmp)^4

{(\ref{L1MKL})}.
Set
${\lambda_1^{(n)}}= \psi_s \eta(t) \xi_n(\lambdatmp)$, ${\lambda_2^{(n)}}=
{\lambda_1^{(n)}}\lambdatmp^{{1}/{2}}$,
${\lambda_3^{(n)}}= 0$ where
$
\lambdatmp= d^{{(1-s)}/{(1+s)}} n^{-{1}/{(1+s)}}R_{1,f^*
}^{-{2}/{(1+s)}}.
$
Then
for all $n$ satisfying $\frac{\log(M)}{\sqrt{n}} \leq1$ and
%
%e6 #&#
%
\begin{equation}
\label{eq:recondC2bound} \frac{\tilde{C}_1}{\recond{\tilb_2}^2}
\psi_s \sqrt{n} \xi
_n(\lambdatmp)^2 d \leq1,
\end{equation}
the generalization error of $L_1$-MKL is bounded as
%
%e7 #&#
%
\begin{eqnarray}\label{eq:L1bound}
&&\bigl\|\fhat- f^*\bigr\|_{L_2(\Pi)}^2 \leq\frac{\tilde{C}_2}{\recond{\tilb
_2}^2} \biggl(
d^{{(1-s)}/{(1+s)}}n^{-{1}/{(1+s)}}R_{1,f^*}^{{2s}/{(1+s)}}
\nonumber
\\[-8pt]
\\[-8pt]
\nonumber
&&\hspace*{103pt}{} +
d^{{(s-1)}/{(1+s)}} n^{-{1}/{(1+s)}}R_{1,f^*}^{
{2}/{(1+s)}} +
\frac{d \log(M)}{n} \biggr) \eta(t)^2,
\end{eqnarray}
with probability
$1- \exp(- t) -
\exp(-\min\{\frac{\recond{\tilb_2}^4 \log(M)}{\tilde
{C}_1^2 \psi_s^2 n\xi_n(\lambdatmp)^4 d^2},
\frac{\recond{\tilb_2}^2}{\tilde{C}_1 \psi_s \xi_n(\lambdatmp)^2
d} \} )$ for all \mbox{$t \geq1$}.
\end{Theorem}

The proof of Theorem~\ref{th:TheConvergenceRateMain} is provided in
Section S.3 %\ref{proof:TheConvergenceRateMain}
of the supplementary material [\citet{SUPP:AS:Suzuki+Sugiyama:2013}].
The bounds presented in the theorem can be further simplified under
additional conditions.
To show simplified bounds, we assume that $\recond{\tilb_1}$ and
$\recond{\tilb_2}$ are bounded from below by a positive constant; cf.
the restricted eigenvalue condition, \citet{AS:Bickel+etal:2009}. There
exists $C_2>0$ such that $\recond{\tilb_2} \geq\recond{\tilb_1}
\geq C_2$.
This condition is satisfied if $\recond{16(1+\sqrt{d})} \geq C_2$
because $\frac{ \sqrt{d} \max_{m\in I_0} \|g^*_m \|_{\calH_{m}}
}{R_{2,g^*}} \leq\sqrt{d}$.
Then we obtain simplified bounds with weak conditions.
If $R_{1,f^*} \leq C d$ with a constant $C$ (this holds if $\|
f^*_m\|_{\calH_{m}} \leq C$ for all $m$),
then the first term in the learning rate \eqref{eq:L1bound} of
$L_1$-MKL dominates the second term, and thus equation (\ref{eq:L1bound}) becomes
%
%e8 #&#
%
\begin{equation}
\bigl\|\fhat- f^*\bigr\|_{L_2(\Pi)}^2 \leq O_p \biggl(
d^{{(1-s)}/{(1+s)}}n^{-{1}/{(1+s)}}R_{1,f^*}^{{2s}/{(1+s)}} +
\frac{d \log(M)}{n} \biggr). \label{eq:roughL1bound}
\end{equation}
Similarly, as for the bound of elastic-net MKL,
if $R^2_{2,g^*} \leq C n^{{q}/{(1+s)}} d$ with a constant $C$
(this holds if $\|g^*_m\|_{\calH_{m}} \leq\sqrt{C}$ for all $m$),
then equation (\ref{eq:Elastbound}) becomes
%
%e9 #&#
%
\begin{eqnarray}\label{eq:roughElastbound}
&&\bigl\|\fhat- f^*\bigr\|_{L_2(\Pi)}^2
\nonumber
\\[-8pt]
\\[-8pt]
\nonumber
&&\qquad\leq O_p \biggl(
d^{
{(1+q)}/{(1+q+s)}}n^{-{(1+q)}/{(1+q+s)}}R_{2,g^*}^{{2s}/{(1+q+s)}} +
\frac{d \log(M)}{n} \biggr).
\end{eqnarray}
Here notice that the tail probability can be bounded as
\begin{eqnarray*}
\exp\biggl(-\min\biggl\{\frac{\recond{\tilb_1}^4 \log
(M)}{\tilde
{C}_1^2 \psi_s^2 n\xi_n(\lambdatmp)^4 d^2}, \frac{\recond{\tilb
_1}^2}{\tilde{C}_1 \psi_s \xi_n(\lambdatmp)^2
d} \biggr\}
\biggr) &\leq&\exp\bigl(-\min\bigl\{\log(M),\sqrt{n} \bigr\}
\bigr)
\\
&=& \frac{1}{M},
\end{eqnarray*}
under the conditions of equation (\ref{eq:recondC1bound}) and $\frac{\log(M)}{\sqrt{n}}
\leq1$
[the same inequality also holds under equation (\ref{eq:recondC2bound}),
even if we replace $\recond{\tilb_1}$ with $\recond{\tilb_2}$].

%!!!!!!!!!!!!!!!!!!!!!!!!!!!!!!!!!!!!!!!!!!
%!Sufficient conditions on n.
% Next we examine the conditions \Eqref{eq:recondC1bound} and
% By substituting the value of $\lambdatmp$ determined in the theorem,
% \begin{eqnarray}
% \frac{\tilde{C}_1}{\recond{\tilb_1}^2} \psi_s\max\left(\frac{d^{
% \frac{d^{\frac{q+s}{1+q+s}}}{n^{\frac{1+(3-s)q -
%s^2}{2(1+s)(1+q+s)}}}R_{2,\gstar}^{\frac{2}{1+q+s}},
% d\frac{\log(M)}{\sqrt{n}}\right) \leq1,
% \label{eq:recondC1bound_mod}
% \end{eqnarray}
% and \Eqref{eq:recondC2bound} is equivalent to
% \begin{eqnarray}
% \frac{\tilde{C}_1}{\recond{\tilb_2}^2} \psi_s
% \max\left(\frac{d^{\frac{1+s^2}{1+s}}}{n^{\frac{1-s}{2(1+s)}}}R_{1,
% \frac{d^{\frac{2s}{1+s}}}{n^{\frac{1- s}{2(1+s)}}}R_{1,\fstar}^{
% d\frac{\log(M)}{\sqrt{n}}\right) \leq1.
% \label{eq:recondC2bound_mod}
% \end{eqnarray}
% If we assume $R_{2,\gstar} \leq\sqrt{d}$ and $R_{1,\fstar} \leq d$
%for simplicity
% (these are true if $\|\fstar_m\hnorm{m},\|\gstar_m\hnorm{m} \leq1$
%for all $m$),
% then the above relations \eqref{eq:recondC1bound_mod},
% $$
% n \gg\max\left(d^{\frac{2(1+q+s)}{1-s+q}}, d^{
% $$
% and
% $$
% n \gg\max\left(d^{\frac{2(1+s)^2}{1-s}}, d^{\frac{4(1+s)}{1-s}},
%d^2 \log(M)^2\right),
% $$
% respectively.

We note that, as $s$ becomes smaller (the RKHSs become simpler), both
learning rates of $L_1$-MKL and elastic-net MKL become faster if
$R_{1,f^*},\break R_{2,g^*}\geq1$.
Although the solutions of both $L_1$-MKL and elastic-net MKL are
derived from the same optimization framework \eqref
{eq:primalElasticMKLnonpara},
there appear to be two convergence rates \eqref{eq:roughL1bound} and
\eqref{eq:roughElastbound} that posses different characteristics
depending on ${\lambda_3^{(n)}}= 0$, or not.
There appears to be no dependency on the smoothness parameter $q$ in
bound \eqref{eq:roughL1bound} of $L_1$-MKL, while bound \eqref
{eq:roughElastbound} of elastic-net MKL depends on $q$.
Let us compare these two learning rates on the two situations: $q=0$
and $q>0$.

(i) ($q=0$). In this situation, the true function $f^*$ is not
smooth and $g^*= f^*$ from the definition of $q$.
% Look at the dependency on $d$.
The terms with respect to $d$ are $d^{{(1-s)}/{(1+s)}}$ for $L_1$-MKL~\eqref{eq:roughL1bound} and $d^{{1}/{(1+s)}}$
for elastic-net MKL~\eqref{eq:roughElastbound}.
Thus, $L_1$-MKL has milder dependency on~$d$.
This might reflect the fact that $L_1$-MKL tends to generate sparser solutions.
Moreover, one can check that the learning rate of $L_1$-MKL~\eqref
{eq:roughL1bound} is better than that of elastic-net MKL~\eqref
{eq:roughElastbound} because
Jensen's inequality $R_{1,f^*} \leq\sqrt{d}R_{2,f^*}$ gives
\[
d^{{(1-s)}/{(1+s)}}n^{-{1}/{(1+s)}}R_{1,f^*}^{{2s}/{(1+s)}} \leq
d^{{1}/{(1+s)}}n^{-{1}/{(1+s)}}R_{2,f^*}^{{2s}/{(1+s)}}.
\]
This suggests that, when the truth is nonsmooth, $L_1$-MKL is preferred.

(ii) ($q>0$).
We see that, as $q$ becomes large (the truth becomes smooth), the
convergence rate of elastic-net MKL becomes faster. % if $R_{2,\gstar}
% Look at the exponent of $n$.
The convergence rate with respect to $n$ in the presented bound is
$n^{-{(1+q)}/{(1+q+s)}}$ for elastic-net MKL that is faster than that
of $L_1$-MKL ($n^{-{1}/{(1+s)}}$).
We suggest that this shows that elastic-net MKL properly captures the
smoothness of the truth $f^*$ using the additional $\ell
_2$-regularization term.
As we observed\break above, we obtained a faster convergence bound of
$L_1$-MKL than that of $L_2$-MKL when $q=0$.
However, if $f^*$ is sufficiently smooth ($g^*$ is small), as
$q$ increases,
there appears ``phase-transition,'' that is, the convergence bound of
elastic-net MKL turns out to be faster than that of $L_1$-MKL\break
[$d^{{(1-s)}/{(1+s)}}n^{-{1}/{(1+s)}}R_{1,f^*}^{
{2s}/{(1+s)}} \geq d^{{(1+q)}/{(1+q+s)}}n^{-
{(1+q)}/{(1+q+s)}}R_{2,g^*}^{{2s}/{(1+q+s)}}$].\break
This might indicate that, when the truth $f^*$ is smooth,
elastic-net MKL is preferred.

An interesting observation here is that
depending on the smoothness $q$ of the truth, the preferred
regularization changes.
%This is due to the trade-off between the sparsity and the smoothness.
Here, we would like to point out that the comparison between $L_1$-MKL
and elastic-net MKL is just based on the upper bounds of the
convergence rates.
Thus there is still the possibility that $L_1$-MKL can also make use of
the smoothness $q$ of the true function to achieve a faster rate.
We will give discussions about this issue in Section~\ref{sec:DiscussionAdaptL1}.

Finally, we give a comprehensive representation of Theorem \ref
{th:TheConvergenceRateMain} that gives a clear correspondence to the
minimax optimal rate given in the next subsection.
%
%co3 #&#
%
\begin{Corollary}%[Convergence Rate of $L_1$-MKL and elastic-net MKL]
\label{cor:ComprehensiveRepresentation}
Suppose the same condition as Theorem~\ref{th:TheConvergenceRateMain}.
Define $\sprime= \frac{s}{1+q}$.
Then there exists constant $\tilde{C}'$ depending on $s,c,L,C_1$
such that the following convergence rates hold:

{(\ref{ElasticnetMKL})}.
If $1 \leq R_{2,g^*}$ and $\|g^*_m\|_{\calH_{m}} \leq C\ (\forall
m\in\Istar)$ with a constant $C$,
then for all $p \geq2$, elastic-net MKL achieves the following
convergence rate:
\[
\bigl\|\fhat- f^*\bigr\|_{L_2(\Pi)}^2 \leq
\frac{\tilde{C}'}{\recond{\tilb_1}^2} \biggl( d^{1 -
{2\sprime}/{(p(1+\sprime))}}n^{-{1}/{(1+\sprime
)}}R_{p,g^*}^{{2\sprime}/{(1+\sprime)}}
+ \frac{d \log(M)}{n} \biggr) \eta(t)^2,
\]
with probability
$1- \exp(- t) - 1/M$ for all $t \geq1$.

{(\ref{L1MKL})}.
If $1 \leq R_{1,f^*}$ and $\|f^*_m\|_{\calH_{m}} \leq C\ (\forall
m\in\Istar)$ with a constant $C$,
then for all $p \geq1$, $L_1$-MKL achieves the following convergence rate:
\[
\bigl\|\fhat- f^*\bigr\|_{L_2(\Pi)}^2 \leq
\frac{\tilde{C}'}{\recond{\tilb_2}^2} \biggl( d^{1 -
{2s}/{(p(1+s))}}n^{-{1}/{(1+s)}}R_{p,f^*}^{
{2s}/{(1+s)}}
+ \frac{d \log(M)}{n} \biggr) \eta(t)^2,
\]
with probability
$1- \exp(- t) - 1/M$ for all $t \geq1$.
\end{Corollary}
\begin{pf}
Due to Jensen's inequality,
we always have $R_{2,g^*} \leq d^{{1}/{2}-{1}/{p}}
R_{p,g^*}$ for $p \geq2$ and $R_{1,f^*} \leq d^{1-
{1}/{p}} R_{p,f^*}$ for $p\geq1$.
Thus we have
\begin{eqnarray*}
d^{{1}/{(1+\sprime)}}n^{-{1}/{(1+\sprime)}}R_{2,g^*
}^{{2\sprime}/{(1+\sprime)}} &\leq&
d^{1 - {2\sprime}/{(p(1+\sprime))}}n^{-{1}/{(1+\sprime
)}}R_{p,g^*}^{{2\sprime}/{(1+\sprime)}},
\\
d^{{(1-s)}/{(1+s)}} n^{-{1}/{(1+s)}}R_{1,f^*}^{
{2s}/{(1+s)}} &\leq&
d^{1-{2s}/{(p(1+s))}} n^{-{1}/{(1+s)}}R_{p,f^*}^{{2 s}/{(1+s)}}.
\end{eqnarray*}
Combining this and the discussions to derive equations (\ref{eq:roughL1bound}) and (\ref{eq:roughElastbound}), we
have the assertion.
\end{pf}
Below, we show that bounds \eqref{eq:roughL1bound} and \eqref
{eq:roughElastbound} achieve the minimax optimal rates on the $\ell
_1$-mixed-norm ball and the $\ell_2$-mixed-norm ball,
respectively.

%s3.2 #&#
\subsection{\texorpdfstring{Minimax learning rate of $\ell_p$-mixed-norm ball}
{Minimax learning rate of lp-mixed-norm ball}}
Here we consider a simple setup to investigate the minimax rate.
First, we assume that the input space $\calX$ is
expressed as $\calX= \tilde{\calX}^M$ for some space $\tilde{\calX}$.
% an $M$ product of a space $\tilde{\calX}$, i.e.,
Second, all the RKHSs $\{\calH_m\}_{m=1}^M$ are induced from the same
RKHS $\tilde{\calH}$ defined on~$\tilde{\calX}$.
Finally, we assume that the marginal distribution $\Pi$ of input is
the product of a probability distribution $Q$, that is, $\Pi= Q^M$.
Thus, an input $x=(\tilde{x}^{(1)},\ldots,\tilde{x}^{(M)}) \in\calX
= \tilde{\calX}^M$ is concatenation of $M$ random variables $\{\tilde
{x}^{(m)}\}_{m=1}^M$
independently and identically distributed from the distribution~$Q$.
Moreover, the function class $\totH$ is assumed to be a class of
functions $f$ such that
$
f(x) = f(\tilde{x}^{(1)},\ldots,\tilde{x}^{(M)}) = \sum_{m=1}^M
f_m(\tilde{x}^{(m)}),
$
where $f_m \in\tilde{\calH}$ for all~$m$.
Without loss of generality, we may suppose that all functions in
$\tilde{\calH}$ are centered:
$
\EE_{\tilde{X} \sim Q}[f(\tilde{X})] = 0\ (\forall f \in\tilde
{\calH}).
$
Furthermore, we assume that the spectrum of the kernel $\tilde{k}$
corresponding to the RKHS $\tilde{\calH}$ decays at the rate of
$-\frac{1}{s}$.
That is,
in addition to Assumption~\ref{ass:specass}, we impose the following
lower bound on the spectrum: % there exists $0 < s < 1$ such that %and
%$c_1,c_2$ such that
There exist $c',c$ $(>0)$ such that
%
%e10 #&#
%
\begin{equation}
c'j^{-{1}/{s}} \leq\mu_{j} \leq c
j^{-{1}/{s}}, \label{eq:strongSpecAss}
\end{equation}
where $\{\mu_{j}\}_{j}$ is the spectrum of the integral operator
$T_{\tilde{k}}$ with respect to the kernel~$\tilde{k}$; see equation (\ref{eq:spectralRepre}).
We also assume that the noise $\{\varepsilon_i\}_{i=1}^n$ is generated by
the Gaussian distribution with mean 0 and standard deviation $\sigma$.

Let $\calH_{0}(d)$ be the set of functions with $d$ nonzero
components in $\totH$
defined by
$
\calH_{0}(d):= \{ (f_1,\ldots,f_M) \in\totH\mid\#\{m \mid\|
f_m\|_{\calH_{m}} \neq0\} \leq d \}.
$
We define the $\ell_p$-mixed-norm ball ($p\geq1$) with radius $R$ in
$\calH_0(d)$ as
\begin{eqnarray*}
&&\calH_{\ell_p}^{d,q}(R):= \Biggl\{ f = \sum
_{m=1}^M f_m \bigg|  \exists(g_1,
\ldots,g_M) \in\calH_0(d), f_m =
T_m^{{q}/{2}} g_m,\\
&&\hspace*{184pt}\Biggl( \sum_{m=1}^M
\|g_m\|_{\calH_{m}}^p \Biggr)^{{1}/{p}} \leq R \Biggr\}.
\end{eqnarray*}
In \citet{JMLR:Raskutti+Martin:2012}, the minimax learning rate on
$\calH_{\ell_{\infty}}^{d,0}(R)$ (i.e., $p=\infty$ and $q=0$) was
derived.\footnote{%
The set $\calF_{M,d,\calH}(R)$ in \citet{JMLR:Raskutti+Martin:2012}
corresponds to $\calH_{\ell_{\infty}}^{d,0}(R)$ in the current paper.}
We show (a lower bound of) the minimax learning rate for more general settings
($1\leq p \leq\infty$ and $0\leq q \leq1$)
in the following theorem.

%th4 #&#
%
\begin{Theorem}
\label{eq:LowerboundOfElastMKL}
Let $\sprime= \frac{s}{1+q}$.
Assume $d \leq M/4$.
Then the minimax learning rates are lower bounded as follows.
If the radius of the $\ell_p$-mixed-norm ball $R_p$ satisfies
$R_p\geq d^{{1}/{p}}\sqrt{\frac{\log(M/d)}{n}}$,
there exists a constant $\widehat{C}_1$ such that
%
%e11 #&#
%
\begin{eqnarray}\label{eq:minimaxLp}
&&\inf_{\fhat} \sup_{f^*\in\calH_{\ell_p}^{d,q}(R_p)}\EE\bigl[\bigl\|
\fhat-
f^*\bigr\|_{L_2(\Pi)}^2\bigr]
\nonumber
\\[-8pt]
\\[-8pt]
\nonumber
&&\qquad \geq\widehat{C}_1
\biggl( d^{1 - {2\sprime}/{(p(1+\sprime))}} n^{-
{1}/{(1+\sprime)}}R_p^{{2 \sprime}/{(1+\sprime)}} +
\frac
{d \log(M/d)}{n} \biggr),
\end{eqnarray}
where ``inf'' is taken over all measurable functions of the samples $\{
(x_i,y_i)\}_{i=1}^n$, and the expectation is taken for the sample distribution.
%Similarly, we have the following minimax rate for $p=\infty$:
%for $R_{\infty}\geq\sqrt{\frac{\log(M/d)}{n}}$.
\end{Theorem}

A proof of Theorem~\ref{eq:LowerboundOfElastMKL} is provided in
Section S.7 %\ref{proof:LowerboundOfElastMKL}
of the supplementary material [\citet{SUPP:AS:Suzuki+Sugiyama:2013}].

Substituting $q=0$ and $p=1$ into the minimax learning rate \eqref
{eq:minimaxLp},
we see that the learning rate \eqref{eq:roughL1bound} of $L_1$-MKL
achieves the minimax optimal rate of the $\ell_1$-mixed-norm ball for $q=0$.
Moreover, the learning rate of $L_1$-MKL (i.e., minimax optimal on the
$\ell_1$-mixed-norm ball)
is \textit{fastest} among all the optimal minimax rates on $\ell
_p$-mixed-norm ball for $p\geq1$ when $q=0$.
To see this,\vspace*{1pt} let $R_{p,f^*}:= (\sum_{m} \|f^*_m \|_{\calH_{m}}^p
)^{{1}/{p}}$;
then, as in the proof of Corollary \ref
{cor:ComprehensiveRepresentation}, we always have $R_{1,f^*} \leq
d^{1-{1}/{p}} R_{p,f^*} \leq d R_{\infty,f^*}$ due to
Jensen's inequality,
and consequently we have
%
%e12 #&#
%
\begin{eqnarray} \label{eq:JensenMinimax}
d^{{(1-s)}/{(1+s)}} n^{-{1}/{(1+s)}}R_{1,f^*}^{{2s}/{(1+s)}}
&\leq&
d^{1-{2s}/{(p(1+s))}} n^{-{1}/{(1+s)}}R_{p,f^*}^{{2
s}/{(1+s)}}
\nonumber
\\[-8pt]
\\[-8pt]
\nonumber
&\leq& d
n^{-{1}/{(1+s)}}R_{\infty,f^*}^{{2 s}/{(1+s)}}.
\end{eqnarray}

On the other hand, the learning rate \eqref{eq:roughElastbound} of
elastic-net MKL achieves the minimax optimal rate \eqref{eq:minimaxLp}
on the $\ell_2$-mixed-norm ball ($p=2$).
When $q=0$, the rate of elastic-net MKL is slower than that of $L_1$-MKL,
but
the optimal rate is achieved over the whole range of smoothness
parameter $0 \leq q \leq1$,
which is advantageous against $L_1$-MKL.
Moreover, the optimal rate on the $\ell_2$-mixed-norm ball is still
faster than that on the $\ell_\infty$-mixed-norm ball due to relation
\eqref{eq:JensenMinimax}.
%It is still faster than the learning rate on $\ell_{
%$$d^{\frac{1}{1+\sprime}} n^{-\frac{1}{1+\sprime}}R_{2,\gstar}^{

The learning rates of both $L_1$ and elastic-net MKL coincide with the
minimax optimal rate of the $\ell_{\infty}$-mixed-norm ball
when the truth is \textit{homogeneous}. For simplicity, assume $q=0$.
If $\|f^*_m\|_{\calH_{m}} = 1$ ($\forall m \in I_0$) and $f^*_m=0$
(otherwise),
then $R_{p,f^*} = d^{{1}/{p}}$. Thus, both rates are
$d n^{-{1}/{(1+s)}} + \frac{d\log(M)}{n}$; that is, the minimax
rate on the $\ell_{\infty}$-mixed-norm ball.
We also notice that this homogeneous situation is the only situation
where those convergence rates coincide with each other.
As we will see later, the existing bounds are the minimax rate on the
$\ell_\infty$-mixed-norm ball and thus are tight only in the
homogeneous setting.

% Now we consider two examples, ``inhomogeneous setting'' and
%``homogeneous setting'', to compare these two bounds:

% (a) $\|\gstar_m\hnorm{m} = m^{-1}$ ($\forall m \in I_0=\{1,\ldots,d
%1$ and $R_{2,\gstar} \leq1$. Thus,
% the learning rate \eqref{eq:roughbound2} of elastic-net MKL and the
%minimax rate on the $\ell_2$-mixed-norm ball are $d^{\frac{1}{1+
% and that on the $\ell_\infty$-mixed-norm ball is $d n^{-\frac{1}{1+
% Therefore, in the first term (the leading term with respect to $n$),
% there is a difference in the $d\qquad^{\frac{\tilde{s}}{1+\tilde{s}}}$
%factor.
% This difference could be $\sqrt{d}$ in the worst case.
% Thus, there appears large discrepancy between the two rates in
%high-dimensional settings. %in the near sparse setting.

% (b)$\|\gstar_m\hnorm{m} = 1$ ($\forall m \in I_0$) (homogeneous
%setting):
% In this situation, $R_{\infty,\gstar} = 1$ and $R_{2,\gstar} =
% $d n^{-\frac{1}{1+\sprime}} + \frac{d\log(M)}{n}$. Here we observe
%that the learning rate \eqref{eq:roughbound2} of elastic-net MKL
%coincides
% with the minimax rate on the $\ell_\infty$-mixed-norm ball.
% We also notice that
% the homogeneous setting is the only situation where those two rates
%coincide with each other.
% As seen later, the existing bounds by previous works are the minimax
%rate on the $\ell_\infty$-mixed-norm ball, thus are tight only in the
%homogeneous setting.

%
%We presented new learning rates for $L_1$-MKL and elastic-net MKL,
%which are faster than the existing bounds of several MKL formulations.
%

%s4 #&#
\section{Optimal parameter selection}
We need the knowledge of parameters such as $q,s,d,R_{1,f^*
},R_{2,g^*}$
to obtain the optimal learning rate shown in Theorem \ref
{th:TheConvergenceRateMain}; %regularization parameters $\lambdaone,
however, this is not realistic in practice.

To overcome this problem, we give an algorithmic procedure such as
\textit{cross-validation} to achieve the optimal learning rate.
Roughly speaking, we split the data into the training set and the
validation set
and utilize the validation set to choose the optimal parameter.
Given the data $D = \{(x_i,y_i)\}_{i=1}^n$,
the training set $\Dtr$ is generated by using the half of the given
data $\Dtr= \{(x_i,y_i)\}_{i=1}^{n'}$ where $n' = \lfloor\frac{n}{2}
\rfloor$
and the remaining data is used as the validation set $\Dte= \{
(x_i,y_i)\}_{i=n'+1}^{n}$.
Let $\fhat_{\Lambda}$ be the estimator given by our MKL formulation
\eqref{eq:primalElasticMKLnonpara}
where the parameter setting $\Lambda= ({\lambda_1^{(n)}},{\lambda_2^{(n)}},{\lambda_3^{(n)}})$ is employed, and
the training set $\Dtr$ is used instead of the whole data set~$D$.

We utilize a \textit{clipped estimator} so that the estimator bounded
in a way that makes the validation procedure effective.
Given the estimator $\fhat_{\Lambda}$ and a positive real $B > 0$,
the clipped estimator $\fcl_{\Lambda}$ is given as
\[
\fcl_{\Lambda}(x):= \cases{ B, &\quad $\bigl(B \leq\fhat_{\Lambda}(x)
\bigr)$, \vspace*{2pt}
\cr
\fhat_{\Lambda}(x),& \quad $\bigl(-B <
\fhat_{\Lambda}(x) < B\bigr),$\vspace*{2pt}
\cr
-B,& \quad $\bigl(
\fhat_{\Lambda}(x) \leq-B\bigr).$}
\]

To appropriately choose $B$,
we assume that
we can roughly estimate
the sup-norm $\|f^*\|_{\infty}$ of the true function, and $B$ is
set to satisfy $\|f^*\|_{\infty} < B$.
This assumption is not unrealistic because if we set $B$ sufficiently
large so that we have $\max_{i}|y_i| < B$,
then with high probability such $B$ satisfies $\|f^*\|_{\infty} <
B $.
%Instead of estimating the range of $y$, we can set $B$ as $\|\fstar_m
%$\|\fstar_m\|_{\infty} + L$ bounds the range of $y$ from above (see
%Assumption~\ref{ass:basic} for the definition of $L$).
%For simplicity, we assume that $B$ is greater than (but proportional
%to) $\|\fstar_m\|_{\infty} + L$.
It should be noted that if $\|f^*\|_{\infty} < B$,
the generalization error of the clipped estimator $\fcl_{\Lambda}$ is not
greater than that of the original estimator~$\fhat_{\Lambda}$,
\[
\bigl\| \fcl_{\Lambda} - f^*\bigr\|_{L_2(\Pi)} \leq\bigl\| \fhat_{\Lambda}
- f^*\bigr\|_{L_2(\Pi)},
\]
because $|\fcl_{\Lambda}(x) - f^*(x)| \leq|\fhat_{\Lambda}(x)
- f^*(x)|$ for all $x \in\calX$.

Now, for a finite set of parameter candidates $\Theta_n \subset\Real
_+ \times\Real_+ \times\Real_+$,
we choose an optimal parameter that minimizes the error on the
validation set,
%
%e13 #&#
%
\begin{equation}
\Lambda_{\Dte}:= \mathop{\argmin}_{\Lambda\in\Theta_n} \frac{1}{|\Dte|}
\sum
_{(x_i,y_i)\in\Dte} \bigl(\fcl_{\Lambda}(x_i) -
y_i\bigr)^2. \label{eq:validationproc}
\end{equation}
Then we can show that the estimator $\fcl_{\Lambda_{\Dte}}$
achieves the optimal learning rate. % as in the following theorem.
To show this, we determine the finite set $\Theta_n$ of the candidate
parameters as follows:
let $\Gamma_n:= \{1/n^2,2/n^2,\ldots,1\}$ and
\begin{eqnarray*}
\Theta_n &=& \bigl\{(\lambda_1,\lambda_2,
\lambda_3)\mid\lambda_1,\lambda_3 \in
\Gamma_n,\lambda_2 = \lambda_1
\lambda_3^{
{1}/{2}} \bigr\}
\\
&&{}\cup\bigl\{(\lambda_1,\lambda_2,\lambda_3)
\mid\lambda_1,\lambda\in\Gamma_n,\lambda_2 =
\lambda_1 \lambda^{{1}/{2}}, \lambda_3=0\bigr\}.
\end{eqnarray*}
With this parameter set, we have the following theorem that shows the
optimality of the validation procedure \eqref{eq:validationproc}.
%
%th5 #&#
%
\begin{Theorem}
\label{th:ValidationOptimal}
Suppose Assumptions~\ref{ass:basic}--\ref{ass:supnorm} are satisfied.
Assume $R_{1,f^*}$, \mbox{$R_{2,g^*} \geq1$,} $\recond{\tilb_2} \geq
\recond{\tilb_1} \geq C_2$
and $\|f^*_m\|_{\calH_{m}},\|g^*_m\|_{\calH_{m}} \leq C_3$ with some
constants $C_2,C_3 > 0$,
and
suppose $n$ satisfies $\frac{\log(M)}{\sqrt{n}} \leq1$ and
\[
\frac{\tilde{C}_1}{\recond{\tilb_1}^2} \psi_s \sqrt{n} \xi_n(
\lambdatmp_{(1)})^2 d \leq1\quad \mbox{and}\quad \frac{\tilde
{C}_1}{\recond{\tilb_2}^2}
\psi_s \sqrt{n} \xi_n(\lambdatmp_{(2)})^2
d \leq1,
\]
where
$
\lambdatmp_{(1)} = d^{{1}/{(1+q+s)}} n^{-
{1}/{(1+q+s)}}R_{2,g^*}^{-{2}/{(1+q+s)}},
$
$\lambdatmp_{(2)} = d^{{(1-s)}/{(1+s)}} n^{-{1}/({1+s)}}\times R_{1,f^*}^{-{2}/{(1+s)}}$
and
$\tilde{C}_1$ is the constant introduced in the statement of Theorem
\ref{th:TheConvergenceRateMain}.
Then %under the same settings as Theorem
there exist a universal constant $\tilde{C}_4$\vadjust{\goodbreak} and a constant $\tilde
{C}_3$ depending on $s,c,L,C_1,C_2,C_3$ such that
\begin{eqnarray*}
&&\bigl\|\fcl_{\Lambda_{\Dte}} - f^*\bigr\|_{L_2(\Pi)}^2
\\
&&\qquad\leq\tilde{C}_3 \biggl( d^{{(1-s)}/{(1+s)}}n^{-
{1}/{(1+s)}}R_{1,f^*}^{
{2s}/{(1+s)}}\\
&&\quad\hspace*{42pt}{}\wedge d^{{(1+q)}/{(1+q+s)}}n^{-{(1+q)}/{(1+q+s)}}R_{2,g^*}^{
{2s}/{(1+q+s)}} +
\frac{d \log(M)}{n} \biggr) \eta(t)^2
\\
&&\quad\qquad{} + \tilde{C}_4 \frac{B^2(\tau+ \log(1+n))}{n},
\end{eqnarray*}
with probabitlity $1-2\exp(-t) - \exp(-\tau) -\frac{2}{M}$,
where $a \wedge b$ means $\min\{a,b\}$.
\end{Theorem}
This can be shown by combining our bound in Theorem \ref
{th:TheConvergenceRateMain}
and the technique used in Theorem 7.2 of \citet{Book:Steinwart:2008}.
According to Theorem~\ref{th:ValidationOptimal},
the estimator $\fcl_{\Lambda_{\Dte}}$ with the validated parameter
$\Lambda_{\Dte}$
achieves the minimum learning rate among
the oracle bound for $L_1$-MKL \eqref{eq:roughL1bound}
and that for elastic-net MKL \eqref{eq:roughElastbound} if $B$ is
sufficiently small.
Therefore, the optimal rate is almost attainable [at the cost of the
term $\frac{B^2 \log(1+n)}{n}$]
by a simple executable algorithm.

%s5 #&#
\section{Comparison with existing bounds}
In this section,
we compare our bound with the existing bounds.
Roughly speaking, the difference between the existing bounds is
summarized in the following two points
(see also Table~\ref{tab:RelationOfBounds} summarizing the relations
between our analysis and existing analyses):
\begin{longlist}[(a)]
\item[(a)] Our learning rate achieves the minimax rate of the $\ell
_1$-mixed-norm ball or the $\ell_2$-mixed-norm ball,
instead of the $\ell_\infty$-mixed-norm ball.
\item[(b)] Our bound includes the smoothing parameter $q$ (Assumption
\ref{ass:convolution}),
and thus is more general and faster than existing bounds.
\end{longlist}

The first bound on the convergence rate of MKL was derived by \citet
{COLT:Koltchinskii:2008},
which assumed $q=1$ and $\frac{1}{d} \sum_{m \in I_0}(\|g^*
_m\|_{\calH_{m}}^2/\break\|f^*_m\|_{\calH_{m}}^2) \leq C$.
Under these rather strong conditions, they showed the bound
\[
d^{{(1-s)}/{(1+s)}}n^{-{1}/{(1+s)}} + \frac{d\log(M)}{n}.
\]
Our convergence rate \eqref{eq:roughL1bound} of $L_1$-MKL achieves
this learning rate \textit{without} the two strong conditions.
Moreover, for the smooth case $q=1$, we have shown that elastic-net MKL
has a faster rate $n^{-{2}/{(2+s)}}$ instead of
$n^{-{1}/{(1+s)}}$ with respect to~$n$.

The second bound was given by \citet{AS:Meier+Geer+Buhlmann:2009},
which shows
\[
\biggl(\frac{\log(M)}{n} \biggr)^{{1}/{(1+s)}} \bigl(d + R_{2,f^*}^2
\bigr)\vadjust{\goodbreak}
\]
for elastic-net regularization under the condition $q=0$.
Their bound almost achieves the minimax rate on the $\ell_\infty
$-mixed-norm ball
except the $\log(M)$ factor.
Compared with our bound \eqref{eq:roughElastbound}, their bound has
the additional $\log(M)$ factor and the term with respect to $d$ and
$R_{2,f^*}$
is larger than $d^{{1}/{(1+s)}} R_{2,f^*}^{{2s}/{(1+s)}}$ in
our learning rate of elastic-net MKL
because Young's inequality yields
\[
d^{{1}/{(1+s)}} R_{2,f^*}^{{2s}/{(1+s)}} \leq\frac{1}{1+s} d +
\frac{s}{1+s} R_{2,f^*}^2 \leq d + R_{2,f^*}^2.
\]
Moreover, our result for elastic-net MKL covers all $0\leq q \leq1$.

Most recently, \citet{AS:Koltchinskii+Yuan:2010} presented the bound
\[
n^{-{1}/{(1+s)}}(d + R_{1,f^*} ) + \frac{d\log(M)}{n}
\]
for $L_1$-MKL and $q=0$.
Their bound achieves %exactly
the minimax rate on the $\ell_\infty$-mixed-norm ball,
but is looser than our bound \eqref{eq:roughL1bound} of $L_1$-MKL
because, by Young's inequality, we have
\[
d^{{(1-s)}/{(1+s)}}R_{1,f^*} ^{{2 s}/{(1+s)}} \leq\frac{1-s}{1+s}
d +
\frac{2 s}{1+s} R_{1,f^*} \leq d + R_{1,f^*}.
\]
In fact, their bound is $d^{{2s}/{(1+s)}}$ times slower than ours if
the ground truth is \textit{inhomogeneous}.
To see this, suppose $\|f^*_m\|_{\calH_{m}} = m^{-1}$ ($m\in I_0 =\{
1,\ldots,d\}$) and \mbox{$f^*_m=0$} (otherwise).
Then their bound is $n^{-{1}/{(1+s)}}d + \frac{d\log(M)}{n}$,
while our bound for $L_1$-MKL is $n^{-{1}/{(1+s)}}d^{
{(1-s)}/{(1+s)}} + \frac{d\log(M)}{n}$.
%Besides the resultant convergence rate,
Moreover,
their formulation of $L_1$-MKL is slightly different from ours.
In their formulation,
there are additional constraints such that $\|f_m\|_{\calH_{m}} \leq
R_m\ (\forall m)$ with some constants $R_m$
in the optimization problem described in equation (\ref{eq:primalElasticMKLnonpara}).
Due to these constraints, their formulation is a bit different from the
practically used one
(in practice, we do not usually impose such constrains).
Instead, our analysis requires an additional assumption on the sup-norm
(Assumption~\ref{ass:supnorm})
to control the discrepancy between the empirical and population means
of the square of an element in RKHS, $\frac{1}{n}\sum
_{i=1}^nf_m^2(x_i) - \EE[f_m^2]\ (f_m \in\calH_m)$.
In addition, they assumed the \textit{global boundedness}; that is,
the sup-norm of $f^*$ is bounded by a constant, $\|f^*\|
_{\infty} = \|\sum_{m=1}^M f^*_m\|_{\infty} \leq C$.
This assumption is standard and does not affect the convergence rate in
single kernel learning settings.
However, in MKL settings, it is pointed out that the rate is not
minimax optimal in large $d$ regime [in particular $d=\Omega(\sqrt
{n})$] under the global boundedness
[\citet{JMLR:Raskutti+Martin:2012}].
Our analysis omits the global boundedness by utilizing the sup-norm
assumption (Assumption~\ref{ass:supnorm}).

All of the bounds explained above focused on either $q=0$ or $1$.
On the other hand, our analysis is more general in that the whole range
of $0\leq q \leq1$ is covered.

%s6 #&#
\section{\texorpdfstring{Discussion about adaptivity of $\ell_1$-regularization}
{Discussion about adaptivity of l1-regularization}}
\label{sec:DiscussionAdaptL1}
In this section, we discuss the issue, ``is it really true that $\ell
_1$-regularization cannot possess adaptivity to the smoothness?''
According to Theorem~\ref{th:TheConvergenceRateMain} and the following
discussion,
the convergence rate of $L_1$-MKL does not have dependency on the
smoothness of the true function.
However, this is just an upper bound.
Thus, there is still possibility that $L_1$-MKL can make use of the
smoothness of the true function.
We give some remarks about this issue.

According to our analysis, it is difficult to improve the bound of
Theorem~\ref{th:TheConvergenceRateMain} without
any additional assumptions.
On the other hand, it is possible to show this if we may assume some
additional conditions.

A technical reason that makes it difficult to show adaptivity of
$L_1$-MKL %to the smoothness
is that the $\ell_1$-regularization is not differentiable at $0$.
Indeed, the sub-gradient of $\|f_m\|_{\calH_{m}}$ is $f_m/\|f_m\|
_{\calH_{m}}$
if $f_m \neq0$,
and compared with that of $\|f_m\|_{\calH_{m}}^2$ (which is $f_m$), there
is a difference of a factor $1/\|f_m\|_{\calH_{m}}$.
This makes it difficult to control the behavior of the estimator around 0.
To avoid this difficulty, we assume that the estimator $\fhat_m$ is
bounded below as follows.

%as5 #&#
%
\begin{Assumption}[(Lower bound assumption)]
\label{ass:fhatlowerbound}
There exist constants $h_m > 0\ (m\in\Istar)$ such that
\renewcommand{\theequation}{A\arabic{Assumption}}
\begin{equation}
 \|\fhat_m\|_{\calH_{m}} \geq
h_m\qquad  (\forall m \in\Istar),
\end{equation}
with probability $1-p_n$.
\end{Assumption}
We will give a justification of this assumption later (Lemma \ref
{lemm:fhatnorm_lowerbound}). If we admit this assumption, we have the
following convergence bound.
%Before showing the statement, define
Define
\begin{eqnarray*}
\hat{R}_{2,g^*} &:=& \biggl(\sum_{m\in\Istar}
\frac{\|g^*
_m\|_{\calH_{m}}^2}{h_m} \biggr)^{{1}/{2}},
\\
b_3 &:= &32 \biggl(1 + \frac{ \sqrt{d} \max_{m\in I_0} (\|g^*_m
\|_{\calH_{m}}/h_m) }{\hat{R}_{2,g^*}} \biggr).
\end{eqnarray*}

%th6 #&#
%
\begin{Theorem}
\label{th:L1smoothBound}
Suppose Assumptions~\ref{ass:basic}--\ref{ass:fhatlowerbound} are
satisfied, and
$\|g^*_m\|_{\calH_{m}} \leq C$ for all $m \in\Istar$.
Set
\[
\lambdatmp= d^{{1}/{(1+q+s)}} n^{-{1}/{(1+q+s)}}\hat
{R}_{2,g^*}^{-{2}/{(1+q+s)}}.
\]
Moreover we set ${\lambda_1^{(n)}}$, ${\lambda_2^{(n)}}$ and
${\lambda_3^{(n)}}$
as ${\lambda_1^{(n)}}= 2 \psi_s \eta(t) \xi_n(\lambdatmp)$,
${\lambda_2^{(n)}}=
\max\{\lambdatmp\eta(t),\break {\lambda_1^{(n)}}\lambdatmp^{
{1}/{2}}\}$,
${\lambda_3^{(n)}}= 0$ where $\psi_s$ is same as Theorem \ref
{th:TheConvergenceRateMain}.
Similarly define ${\lambda_1^{(n)}}(t'),\break{\lambda_2^{(n)}}(t')$
corresponding to
some fixed $t'$,
and $\tilde{\lambda} = ({\lambda_2^{(n)}}(t')/{\lambda_1^{(n)}}(t'))^2$.
Then there exist constants $\tilde{C}_3$, $\tilde{C}_3'$, $\tilde{C}_4$
depending on $s,c,L,C_1,C,b_3,t'$
such that
for all $n$ satisfying $\frac{\log(M)}{\sqrt{n}} \leq1$ and
%
%e14 #&#
%
\renewcommand{\theequation}{\arabic{equation}}
\setcounter{equation}{13}
\begin{equation}
\frac{\tilde{C}_3}{\recond{\tilb_3}^2} \psi_s \sqrt{n} \xi_n^2(
\lambdatmp) d \leq1,\qquad \tilde{C}_3' \psi_s
\sqrt{n}\xi_n^2(\tilde{\lambda}) \tilde{\lambda} d \leq
{\lambda_2^{(n)}}\bigl(t'\bigr), \label{eq:ForLargenL1_2}
\end{equation}
we have that
%
%e15 #&#
%
\begin{eqnarray}
\label{eq:L1_q_depend_bound}&& \bigl\|\fhat-
f^*\bigr\|_{L_2(\Pi)}^2
\nonumber
\\[-8pt]
\\[-8pt]
\nonumber
&&\qquad\leq\frac{\tilde{C}_4}{\recond
{b_3}^2}
\biggl(d^{{(1+q)}/{(1+q+s)}}n^{-{(1+q)}/{(1+q+s)}} \hat
{R}_{2,g^*}^{{2s}/{(1+q+s)}} +
\frac{d\log(M)}{n} \biggr) \eta(t)^2,\hspace*{-35pt}
\end{eqnarray}
with probability $1-\exp(-t) - \exp(-t') - 2/M - p_n$.
\end{Theorem}
The proof of Theorem~\ref{th:L1smoothBound} can be found in Section
S.4 %\ref{sec:proofL1SmoothBound}
of the supplementary material [\citet{SUPP:AS:Suzuki+Sugiyama:2013}].
The theorem shows that with the rather strong assumption (Assumption
\ref{ass:fhatlowerbound}), we can show that
$L_1$-MKL also possesses adaptivity to the smoothness.
Bound \eqref{eq:L1_q_depend_bound} is close to the minimax optimal
rate on the $\ell_2$-mixed-norm ball
where $\hat{R}_{2,g^*}$ appears instead of $R_{2,g^*}$.
Here
we observe that
$h_m$ appears in the denominator in $\hat{R}_{2,g^*}$.
Therefore, for small $h_m$, $\hat{R}_{2,g^*}$ is larger than
$R_{2,g^*}$,
which can make bound \eqref{eq:L1_q_depend_bound} larger than that of
elastic-net MKL.
This is due to the indifferentiability of $\ell_1$-regularization as
explained above.

Next, we give a justification of Assumption~\ref{ass:fhatlowerbound}.
%
%le7 #&#
%
\begin{Lemma}
\label{lemm:fhatnorm_lowerbound}
If $\|\fhat_m - f^*_m\|_{L_2(\Pi)} \to0$ in probability, then
\[
P \biggl(\|\fhat_m\|_{\calH_{m}} \geq\frac{\|f^*_m\|_{\calH
_{m}}}{2} \biggr) \to1.
\]
\end{Lemma}
\begin{pf}
On the basis of decomposition \eqref{eq:spectralRepre} of the kernel function,
we write $f^*_m = \sum_{j=1}^\infty a_{j,m} \phi_{j,m}$ and
$\fhat_m = \sum_{j=1}^\infty\hat{a}_{j,m} \phi_{j,m}$.
Then we have that
$
\|f^*_m\|_{\calH_{m}}^2 = \sum_{j=1}^\infty\mu_{j,m}^{-1} a_{j,m}^2.
$
Now we define $J_{f^*_m}$ to be a finite number such that
$\sqrt{\sum_{j=1}^{J_{f^*_m}} \mu_{j,m}^{-1} a_{j,m}^2} \geq
\frac{3}{4}\|f^*_m\|_{\calH_{m}}$.
Noticing that
$
o_p(1) \geq\|\fhat_m - f^*_m\|_{L_2(\Pi)}^2 =  \sum_{j=1}^\infty
(a_{j,m} - \hat{a}_{j,m})^2 \geq\sum_{j=1}^{J_{f^*_m}} (a_{j,m}
- \hat{a}_{j,m})^2,
$
we have that
\begin{eqnarray*}
\|\fhat_m\|_{\calH_{m}} & =& \sqrt{ \sum
_{j=1}^{J_{f^*_m}} \mu_{j,m}^{-1} \hat
{a}_{j,m}^2 + \sum_{j=J_{f^*_m}+1}^{\infty}
\mu_{j,m}^{-1} \hat{a}_{j,m}^2}
\\
& \geq&\sqrt{ \sum_{j=1}^{J_{f^*_m}}
\mu_{j,m}^{-1} \hat{a}_{j,m}^2}
\\
& \geq&\sqrt{ \sum_{j=1}^{J_{f^*_m}}
\mu_{j,m}^{-1} a_{j,m}^2} - \sqrt
{ \sum_{j=1}^{J_{f^*_m}}
\mu_{j,m}^{-1}(a_{j,m}- \hat{a}_{j,m})^2}
\\
& \geq&\frac{3}{4}\bigl\|f^*_m\bigr\|_{\calH_{m}} -
\mu_{J_{f^*
_m}}^{-{1}/{2}} \sqrt{ \sum
_{j=1}^{J_{f^*_m}} (a_{j,m}- \hat
{a}_{j,m})^2} = \frac{3}{4}\bigl\|f^*_m
\bigr\|_{\calH_{m}} - o_p(1).
\end{eqnarray*}
This gives the assertion.
\end{pf}
One can see from the proof that the convergence rate in Lemma \ref
{lemm:fhatnorm_lowerbound} depends on $f^*_m$.
If $d$ is sufficiently small, we observe that the proof of Theorem \ref
{th:TheConvergenceRateMain} gives that
$\|f^*_m - \fhat_m\|_{L_2(\Pi)} \stackrel{p}{\to} 0$ ($m\in\Istar$).
In this situation,
if we set $h_m = \|f^*_m\|_{\calH_{m}}/2$, $\|f^*_m\|_{\calH_{m}}
\geq
h_m\ (m\in\Istar)$ is satisfied with high probability for sufficiently
large $n$.

The above discussion seems a proper justification to support the
adaptivity of $\ell_1$-regularization.
However, we would like to remark the following two concerns about the
discussion.
First, in a situation where $d$ increases as the number of samples increases,
it is hardly expected that $\|f^*_m\|_{\calH_{m}} > c$ with~some
positive constant~$c$.
It is more natural to suppose that\break $\min_{m\in\Istar}\|f^*
_m\|_{\calH_{m}} \to0$ as $d$ increases.
In that situation, $\hat{R}_{2,g^*}$ becomes much larger as $d$ increases.
Second, since $T_m$ is not invertible, $\|g^*_m\|_{\calH_{m}}/\|f^*
_m\|_{\calH_{m}}$ is not bounded.
Thus for $h_m = \|f^*_m\|_{\calH_{m}}/2$, we have no guarantee that
$\hat{R}_{2,g^*}$ is reasonably small so that the convergence
bound \eqref{eq:L1_q_depend_bound} is meaningful.
Both of these two concerns are caused by the indiffirentiability of
$\ell_1$-regularization at $0$.
Moreover these concerns are specific to high-dimensional situations.
If $d=M=1$ (or $d$ and $M$ are sufficiently small),
then we do not need to worry about such issues.
%We need more sophisticated technique to deal with the behavior around
%$0$.
%We conjecture that.

We have shown that in a restrictive situation, $\ell_1$-regularization
can possess adaptivity to the smoothness of the true function
and achieve a near minimax optimal rate on the $\ell_2$-mixed-norm ball.
It is a future work to clarify whether the lower bounded assumption
(Assumption~\ref{ass:fhatlowerbound}) is a necessary condition or not.

%
%s7 #&#
\section{Conclusion}
We have presented a new learning rate of both $L_1$-MKL and elastic-net MKL,
which is tighter than the existing bounds of several MKL formulations.
According to our bound,
the learning rates of $L_1$-MKL and elastic-net MKL achieve the minimax
optimal rates on the $\ell_1$-mixed-norm ball and the $\ell
_2$-mixed-norm ball, respectively,
instead of the $\ell_\infty$-mixed-norm ball.
%Our bound includes a parameter $s$ representing the complexity of the
%RKHSs and another parameter $q$ controlling the smoothness of the
%truth.
We have also shown that a procedure like cross validation gives the
optimal choice of the parameters.
%We observed that,
%depending on the smoothness of the ground truth, the preferred method
%($L_1$-MKL or elastic-net MKL) changes.
We have discussed a relation between the regularization and the
convergence rate.
Our theoretical analysis suggests that there is a trade-off between the
sparsity and the smoothness;
that is, if the true function is sufficiently smooth, elastic-net
regularization is preferred; otherwise,
$\ell_1$-regularization is preferred.
This theoretical insight supports the recent experimental results
[\citet
{UAI:Cortes+etal:2009},
\citet{NIPS:Marius+etal:2009},
\citet{NIPSWS:ElastMKL:2009}] such
that intermediate regularization between $\ell_1$ and $\ell_2$ often
shows favorable performances.

\begin{appendix}
%s8 #&#
\section{Evaluation of entropy number}
\label{appendix:CoveringNumber}
Here, we give a detailed characterization of the covering number in
terms of the spectrum using the operator $T_m$.
Accordingly, we give the complexity of the set of
%$f$'s
functions
satisfying the convolution assumption (Assumption~\ref{ass:convolution}).
We extend the domain and the range of the operator $T_m$ to the whole
space of $L_2(\Pi)$ and define its power $T_m^{\beta}\dvtx L_2(\Pi
)\to L_2(\Pi)$ for
$\beta\in[0,1]$ as
\[
T_m^\beta f:= \sum_{k=1}^{\infty}
\mu_{k,m}^\beta\langle f, \phi_{k,m}
\rangle_{L_2(\Pi)} \phi_{k,m}\qquad \bigl(f\in L_2(\Pi)\bigr).
\]
Moreover, we define a Hilbert space $\calH_{m,\beta}$ as
\[
\calH_{m,\beta}:= \Biggl\{\sum_{k=1}^{\infty}
b_k \phi_{k,m} \Biggm|\sum_{k=1}^\infty
\mu_{k,m}^{-\beta} b_k^2 < \infty\Biggr
\},
\]
and equip this space with the Hilbert space norm
$
\llVert\sum_{k=1}^{\infty} b_k \phi_{k,m} \rrVert_{\calH
_{m,\beta}}:=  \sqrt{\sum_{k=1}^\infty\mu_{k,m}^{-\beta} b_k^2}.
$
One can check that $\calH_{m,1} = \calH_m$; see Theorem 4.51 of
\citet
{Book:Steinwart:2008}.
Here we define, for $R>0$,
%
%e16 #&#
%
\begin{equation}
\calH_{m}^{q}(R):= \bigl\{f_m =
T_m^{{q}/{2}}g_m \mid g_m\in\calH
_m, \|g_m\|_{\calH_{m}} \leq R \bigr\}. \label{eq:defHmqR}
\end{equation}
Then we obtain the following lemma.
%
%le8 #&#
%
\begin{Lemma}
\label{lemm:HmqEquiv}
$\calH_{m}^{q}(1)$ is equivalent to the unit ball of $\calH_{m,1+q}$:
$\calH_{m}^{q}(1) = \{ f_m \in\calH_{m,1+q} \mid\|f_m \|_{\calH
_{m,1+q}} \leq1 \}$.
\end{Lemma}
This can be shown as follows. For all $f_m \in\calH_{m}^{q}(1)$,
there exists $g_m \in\calH_m$ such that $f_m = T_m^{{q}/{2}}g_m$
and $\|g_m \|_{\calH_{m}}\leq1$.
Thus
$g_m = (T_m^{{q}/{2}})^{-1} f_m = \sum_{k=1}^{\infty} \mu
_{k,m}^{-{q}/{2}} \langle f_m, \phi_{k,m}\rangle_{L_2(\Pi)}\phi
_{k,m}$ and
$1 \geq\|g_m\|_{\calH_{m}} =
\sum_{k=1}^{\infty} \mu_{k,m}^{-1} \langle g_m,\break  \phi_{k,m}\rangle
_{L_2(\Pi)}^2 = \sum_{k=1}^{\infty} \mu_{k,m}^{-(1+q)} \langle
f_m,
\phi_{k,m}\rangle_{L_2(\Pi)}^2$.
Therefore,\vspace*{1pt} $f_m$ is in $\calH_{m}^{q}(1)$ if and only if the norm of
$f$ in $\calH_{m,1+q}$ is well-defined and not greater than 1.

Now Theorem 15 of \citet{COLT:Steinwart+etal:2009} gives an upper bound
of the entropy number of $\calH_{m,\beta}$ as %of the unit ball $
\[
e_i\bigl(\calH_{m,\beta} \to L_2(\Pi)\bigr) \leq C i^{- {\beta}/{(2s)}},
\]
where $C$ is a constant depending on $c,s,\beta$.
This inequality with $\beta= 1$ corresponds to equation 3. %
Moreover, substituting\vadjust{\goodbreak} $\beta=1+q$ into the above equation,
we have
% Lemma~\ref{lemm:HmqEquiv} gives
%
%e17 #&#
%
\begin{equation}
\label{eq:coveringconditionHq} e_i\bigl(\calH_{m,\beta} \to L_2(\Pi)\bigr)
\leq C i^{- {(1+q)}/{(2s)}}. %\log\calN(\varepsilon,\calH_{m}^{q}(1),
\end{equation}
%
%This inequality is utilized to show the minimax optimal rate.

%s9 #&#
\section{\texorpdfstring{Proof of Lemma \lowercase{\protect\ref{lem:incoherence_ineq}}}{Proof of Lemma 1}}
\label{sec:appendixLemm}
\begin{pf*}{Proof of Lemma~\ref{lem:incoherence_ineq}}
For $J = I^c$, we have
\begin{eqnarray*}
P f^2 &=&\|f_I \|_{L_2(\Pi)}^2 + 2
\langle f_I, f_J \rangle_{L_2(\Pi)} +
\|f_J\|_{L_2(\Pi)}^2
\nonumber
\\
& \geq&\|f_I \|_{L_2(\Pi)}^2 - 2 \rho(I) \|
f_I\|_{L_2(\Pi)} \| f_J \|_{L_2(\Pi)} +
\|f_J \|_{L_2(\Pi)}^2
\nonumber
\\
& \geq&\bigl(1- \rho(I)^2\bigr) \| f_I
\|_{L_2(\Pi)}^2 \geq\bigl(1- \rho(I)^2\bigr)
\kmin(I) \biggl(\sum_{m\in I}\| f_m
\|_{L_2(\Pi)
}^2 \biggr), %\label{eq:firstboundforbasic}
\end{eqnarray*}
where we used Cauchy--Schwarz's inequality in the last line.
% but one.
\end{pf*}
\end{appendix}

\section*{Acknowledgments}
The authors would like to thank Ryota Tomioka, Alexandre B. Tsybakov,
Martin Wainwright and Garvesh Raskutti for suggestive discussions.

\begin{supplement}[id=suppA]
\stitle{Supplementary material for: Fast learning rate of multiple
kernel learning: trade-off between sparsity and smoothness\\}
\slink[doi]{10.1214/13-AOS1095SUPP} %[doi,text={...}] - jei reikian
%suskaldyti doi
\sdatatype{.pdf}
\sfilename{aos1095\_supp.pdf}
\sdescription{Due to space constraints, we have moved the proof of the
main theorem to a supplementary document [\citet
{SUPP:AS:Suzuki+Sugiyama:2013}].}
\end{supplement}

% imsref loaded by akundreckaite, 2013-06-12 09:47:10
%

\printaddresses

\end{document}